\documentclass[%
superscriptaddress,
twocolumn,
showkeys, preprintnumbers, nofootinbib, nobibnotes, 
amsmath,amssymb, aps, pre,
numerical
]{revtex4-1}

\usepackage[dvipsnames]{xcolor}
\usepackage{dynlearn}
\usepackage{mathrsfs}
\usepackage{csquotes} 
\usepackage{upgreek}
\usepackage{silence}
\usepackage{balance}

\SetSymbolFont{stmry}{bold}{U}{stmry}{m}{n} 
\newcommand{\integers}              { \mathbb{Z} }
\newcommand{\nats}                  { \mathbb{N} }
\newcommand{\natsWithZero}          { \nats_0 }
\newcommand{\reals}                 { \mathbb{R} }

\newcommand{\Identity}[1]           { I_{#1} }

\newcommand{\innerProductMath}[2]   { \left< #1 , #2 \right> }

\newcommand{\measure}               { \mu }
\newcommand{\measureAlt}            { \nu }
\newcommand{\sigmaAlgebra}          { \Sigma }

\newcommand{\ObvsAlphabet}          { \mathcal{\Obvs} }
\newcommand{\Obvs}                  { Z }

\newcommand{\obvs}                  { z }

\newcommand{\ObvsPast}              { X }
\newcommand{\obvspast}              { x }

\newcommand{\ObvsPastSpace}         { \bm{\mathcal{\ObvsPast}} }
 
\newcommand{\ObvsFuture}            { Y }
\newcommand{\obvsfuture}            { y }
\newcommand{\ObvsFutureSpace}       { \bm{\mathcal{\ObvsFuture}} }
 
\newcommand{\kernel}                { k }
\newcommand{\kernelPast}            { \kernel^{\ObvsPast} }
\newcommand{\kernelFuture}          { \kernel^{\ObvsFuture} }
\newcommand{\HilbertSpace}          { \mathcal{H} }
\newcommand{\HilbertSpacePast}      { \HilbertSpace^{\ObvsPast} }
\newcommand{\HilbertSpaceFuture}    { \HilbertSpace^{\ObvsFuture} }
 
\newcommand{\condCausalMeasure}[1]  { \measure_{\EquiFunction{#1}} }
\newcommand{\kcausalstate}[1]       { \causalstate_{\kernel\ifx\relax#1\relax\else, {#1}\fi} }
\newcommand{\kCausalState}          { \CausalState_{\kernel} }
\newcommand{\kCausalStateSet}       { \CausalStateSet_{\kernel} }

\newcommand{\ekcausalstate}[1]      { \widehat{\causalstate}{\ifx\relax#1\relax\else_{#1}\fi} }

\newcommand{\ekCausalStateSet}      { \widehat{\CausalStateSet} }

\newcommand{\keM}                   {kernel \eM}
\newcommand{\kcs}                   {kernel causal state\xspace}
\newcommand{\kcss}                  {kernel causal states\xspace}
\newcommand{\ekcs}                  {empirical \kcs}
\newcommand{\ekcss}                 {empirical \kcss}

\newcommand{\coords}                {diffusion coordinates\xspace}

\newcommand{\cdcs}                  {causal diffusion components\xspace}

\newcommand{\tempIndices}           { \tau }
\newcommand{\shiftOperator}         { \uptau }
\newcommand{\CausalStateSpan}       { \mathbb{S} }
\newcommand{\EquiFunction}[1]       { \epsilon \left[ #1\right] }

\newcommand{\SequenceSpace}         { \ObvsAlphabet^{\integers} }
\newcommand{\halfSequenceSpace}     { \ObvsAlphabet^{\nats} }
\newcommand{\halfSequenceSpaceZero} { \ObvsAlphabet^{\natsWithZero} }
\newcommand{\ProbSequenceSpace}     { \mathbb{P} \left( \halfSequenceSpace \right) }

\newcommand{\measurementsymbol}     { m }
\newcommand{\numdatasources}        { D }
\newcommand{\datasource}            { d }
\newcommand{\historylength}         { L }
\newcommand{\historylengthpast}     { \historylength_p }
\newcommand{\historylengthfuture}   { \historylength_f }
\newcommand{\numObvs}               { N }
\newcommand{\totalLength}           { T }

\newcommand{\kernelvec}             { \bm{K} }
\newcommand{\condEmbedCoeff}        { \alpha }
\newcommand{\regularizationValue}   { \gamma }

\newcommand{\gramMat}               { G }
\newcommand{\gramMatpast}           { \gramMat^{\ObvsPast} }

\newcommand{\gramMatCS}             { \gramMat^{\kCausalState} }

\newcommand{\diffEigenvalue}        { \lambda }
\newcommand{\diffREigenvector}      { \psi }
\newcommand{\diffLEigenvector}      { \phi }
\newcommand{\diffTransMat}          { M^{\kCausalState} }
\newcommand{\numDiffCoords}         { M }
\newcommand{\numDegreesFreedom}     { \numDiffCoords }
\newcommand{\diffcoord}[2]          { \diffREigenvector_{{#1}\ifx\relax#2\relax\else, {#2}\fi} }
\newcommand{\diffDist}              { d }

\newcommand{\pendangle}             { \theta }
\newcommand{\pendcoord}             { q }
\newcommand{\pendCSParam}           { \phi }
\newcommand{\totalEnergy}           { E }
\newcommand{\period}                { T }
\newcommand{\periodAvg}             { \overline{\period} }
\newcommand{\equilibriumEnergy}     { \totalEnergy_{\text{sep}} }

\newcommand{\dihedralangle}         { \phi }

\newcommand{\kernelBandwidth}       { \xi }

\newcommand{\obsfun}                { E }
\newcommand{\gapLength}             { G }
\newcommand{\gapfillingTransOp}     { T }

\begin{document}

\title{Inferring Kernel \texorpdfstring{{$\epsilon$}}{e}-Machines:\\
Discovering Structure in Complex Systems}

\author{Alexandra M. Jurgens}
\email{alexandra.jurgens@inria.fr}
\affiliation{INRIA Bordeaux Sud Ouest, 33405 Talence Cedex, France}

\author{Nicolas Brodu}
\email{nicolas.brodu@inria.fr}
\thanks{corresponding author} 
\affiliation{INRIA Bordeaux Sud Ouest, 33405 Talence Cedex, France}

\date{\today}
\bibliographystyle{unsrt}

\begin{abstract}

Previously, we showed that computational mechanic's causal
states---predictively-equivalent trajectory classes for a stochastic dynamical
system---can be cast into a reproducing kernel Hilbert space. The result is a
widely-applicable method that infers causal structure directly from very
different kinds of observations and systems. Here, we expand this method to
explicitly introduce the causal diffusion components it produces. These encode
the kernel causal-state estimates as a set of coordinates in a reduced dimension
space. We show how each component extracts predictive features from data and
demonstrate their application on four examples: first, a simple pendulum---an
exactly solvable system; second, a molecular-dynamic trajectory of
$n$-butane---a high-dimensional system with a well-studied energy landscape;
third, the monthly sunspot sequence---the longest-running available time series
of direct observations; and fourth, multi-year observations of an active crop
field---a set of heterogeneous observations of the same ecosystem taken for over
a decade. In this way, we demonstrate that the empirical kernel causal-states
algorithm robustly discovers predictive structures for systems with widely
varying dimensionality and stochasticity.

\end{abstract}

\maketitle
\tableofcontents


\section{Introduction}
\label{sec:Introduction}

In a broad sense the project of the natural sciences turns on the desire to
uncover hidden structure in physical systems. But what, precisely, is meant by
``hidden structure''? To be sure, we do not mean the most obvious form of
structural analysis as performed in the engineering sciences. Rather, we wish to
access something more abstract---a representation of our system that makes
explicit the global patterns of behavior, and/or the patterns of behavior of
distinguishable system components and the nature of how those components are
related and organized.

Early work in nonlinear dynamical systems developed the field of attractor
reconstruction from data using techniques like delay embeddings \cite{Pack80,
sauer1991embedology}. In this kind of structural analysis we analyze the
topology and geometry of the reconstructed attractor to better understand the
dynamics of the system under study. For example, a system may have a set of
behaviors (or patterns of behavior) it tends towards over time, visualized as
trajectories approaching attracting points or limit cycles in phase space.
Modern dynamical system reconstruction techniques are descendants of this kind
of structural analysis \cite{tan2023selecting, mangiarotti2019gpom}.

If we suppose the observed data are the composition of an observation function,
applied to some underlying dynamical system at each instant of time, then we can
also study how the observations change as the action of the underlying dynamics.
This is the definition of the Koopman operator, acting on the observation
function. Interestingly, this operator is linear, even though the original
dynamical system is not. Given the advantages of linear algebra, a consequent
body of research \cite{budivsic2012koopmanism, brunton2021modern} focuses on how
to best express these operators. Algorithms exist to estimate them, provably
consistently so in the limit of infinite data \cite{klus2020operators,
schmid2022dmd}. This is generally accomplished by finding invariant subspaces
using spectral decomposition approximations \cite{giannakis2019data},
effectively turning complicated dynamics into a simpler linear operations.
However, the price to pay lies in the spectral basis functions, which become
complex objects. These techniques can be extraordinarily powerful and flexible
but struggle when it comes to the interpretability of these discovered latent
spaces, which may be unintuitive. Alternatively, methods exist to model the
non-linear evolution of the dynamical system. Recovering equations of motion
from data is an ancient topic \cite{crutchfield1987equations}, but which is
still active \cite{baddoo2022kernel, brunton2016discovering, gauthier2021ngrc,
mangiarotti2019gpom}. The methods more closely related to our approach need to
account for a stochastic component \cite{Frie08a,wang2022data}.

A complementary perspective is offered by information-theoretic definitions of
structure, which strive to describe systems by formalizing concepts like
\emph{irreducible information} and \emph{complexity}. One such framework is
offered by computational mechanics \cite{crut12a} which casts a system as a
communication channel from its past behavior to its future behavior.
Computational mechanics introduces the \emph{\eM}---a model of a system built
from the \emph{causal states}, the set of predictively-equivalent distributions
over the future given the past. In this setting, structural analysis of the
system corresponds to describing the nature of its causal structure---plainly,
what is the stochastic mapping between the system's past and future? We can
easily construct a system where the answer is none at all---a coin flip for
example. We can also imagine a system where every possible past uniquely
determines the future. In general, we assume real systems lie somewhere in
between these extremes.

Recently, computational mechanics was extended to marry these two styles of
structural analysis in a novel way: a system's causal states may be
consistently and uniquely embedded as points in a Hilbert space
\cite{loomis2023topology}. This allows us to construct the \emph{kernel \eM}, a
new geometric representation of the \eM wherein the causal states are points in
a Hilbert space and the dynamics of the \eM may be represented by a stochastic
model of their evolution. The previous contribution in this series introduced
not only the kernel \eM but also a tractable algorithm for estimating \ekcss
from nearly arbitrary data \cite{brodu22rkhs}. The net result is that
computational mechanics was extended to nearly arbitrary data types, including
real data and with minimal constraints. The second major result was that we now
have a presentation of the causal states imbued with not only
information-theoretic notions of structure, but also geometric ones. This
facilitates applying new, intuitive structural analysis of the causal states,
including manifold learning techniques.
This paper applies these techniques and the \ekcs algorithm to real-valued
data for the first time.

We note that the principles we outline here, of describing and predicting the
state-space structure of a complex system, is a well-developed tradition in
time-series forecasting methods and the study of chaos in dynamical systems
\cite{Spro03a} and which is closely related to the subfield of pattern formation
\cite{Cros09a,Cros93a,rupe2024principles}. This said, the prediction of
trajectories in state space is a separate topic. Compared to the methods
mentioned above for recovering dynamical systems equations, in our case the
consistency of the evolution in both data and causal states spaces also needs to
be ensured. We address these issues in a sequel.

\Cref{sec:computuationalmechanics} reviews the
necessary aspects of computational mechanics theory. \Cref{sec:kerneleMachines}
then reviews the theoretical construction of \kcss, as well as the \ekcs
algorithm as introduced previously and the use of a diffusion mapping
technique for nonlinear dimension reduction. Finally, we present the results of
the \ekcs algorithm applied to two simulated examples in
\cref{sec:simulatedexamples} and two real data examples in
\cref{sec:realdatapps}. 

\section{Computational mechanics}
\label{sec:computuationalmechanics}

Traditionally, the basic objects under study in computational mechanics are
conditional distributions over realizations of stochastic processes. In
particular, we generally are interested in building the \emph{\eM}: a
stochastic process's minimal optimally predictive model
\cite{shalizi2001computational}. The states of the \eM are called the
\emph{causal states}, which are sets of past observations of a process that
induce the same distribution over futures.

We take a \emph{stochastic process} $\Process$ to consist of a
$\tempIndices$-indexed random variable $\Obvs$ defined on the measurable space
$\left( \ObvsAlphabet^{\tempIndices}, \sigmaAlgebra, \measure \right)$.
Specific realizations of random variables are denoted by use of the lower case
and time indexing is done by the use of subscripts. For example, we write
$\Obvs_t = \obvs_t$ to say that $\obvs \in \ObvsAlphabet$ is the specific value
of $\Obvs$ at time $t$. The dynamic of the stochastic process is given by the
shift operator, also called the \emph{translation operator}, which is an operator
$\shiftOperator : \ObvsAlphabet^{\tempIndices} \to \ObvsAlphabet^{\tempIndices}$
that maps $t$ to $t+1$: $\shiftOperator \obvs_t = \obvs_{t+1}$. It also acts on
the measure: $ (\shiftOperator \measure) (E) = \measure (\shiftOperator^{-1} E )
$ for $E \in \sigmaAlgebra$. Temporal blocks of the process are given by
$\left\{ \obvs_{a < t \leq b}  : a < b; a, b \in \tempIndices \right\}$.

Computational mechanics is primarily concerned with the prediction of the future
of a system given observations of its past. To be technical about these terms,
we refer to the observation at $t = 0$ as the \emph{present}. We call $\Obvs_{t
\leq 0 }$ the \emph{past} $\ObvsPast$ and $\Obvs_{t > 0}$ the \emph{future}
$\ObvsFuture$. Using distinct symbols for the past and future is both for
readability and to match notation from our previous work \cite{brodu22rkhs}. We
assume that the past and future are defined on their own measurable spaces:
$\left( \ObvsAlphabet^{\tempIndices \leq 0}, \sigmaAlgebra_{\ObvsPast},
\measure_{\ObvsPast} \right)$ and $\left(  \ObvsAlphabet^{\tempIndices > 0},
\sigmaAlgebra_{\ObvsFuture}, \measure_{\ObvsFuture} \right)$, respectively. A
time subscript on a past or future represents the time at which that past ends
or that future beings, as appropriate. So, a specific realization of the past at
time $\tau$ is written $\obvspast_{\tau} = \obvs_{t \leq \tau}$. A corresponding
specific realization of the future is $\obvsfuture_{\tau} = \obvs_{t > \tau}$.
Thus, at each time step an instance of the process $\obvs$ splits into paired
instances of a past and a future: $\obvs = \left( \obvspast_{\tau},
\obvsfuture_{\tau} \right) \, \forall \, \tau \in \tempIndices$.

For the theoretical development of the causal states we make two assumptions
about the stochastic process. We assume the process $\Process$ is conditionally
stationary, which is to say that the conditional distribution over futures given
a specific past $\Pr \left( \ObvsFuture \mid \ObvsPast = \obvspast \right)$ is
the same for all $t \in \integers$ up to a null-measure set. For this reason we
may drop the time index $t$ when it is unnecessary. Conditional stationarity is
a weaker condition than requiring $\Process$ be stationary, which would require
$\shiftOperator \measure = \measure$. We also typically assume that the measure
is ergodic, which is to say that all shift-invariant sets $I \in \SequenceSpace$
are either $\measure(I) = 1$ or $\measure(I) = 0$. Practically speaking, the
\ekcs algorithm as described in \cref{sec:CausalStateEmbeddingAlgo} and
\cref{sec:diffMapping} may be applied to data even when these conditions are not
met (or unable to be tested for), but the strict guarantees of convergence of
the causal states may not hold. Since real-world applications are already
working with finite data and other practical limitations, this is not typically
of great concern. 

We will, however, enforce two very minor constraints. First, the temporal
indices $\tempIndices$ are taken to be the integers $\integers$, which is to say
that the process is \emph{discrete-time}. Second, $\ObvsAlphabet$ is taken to be
compact---typically, either a discrete finite set or a closed interval in
$\reals$ (or a Cartesian product of a finite number of these, which is
sufficient to ensure that $\ObvsAlphabet$ is compact by Tychonoff's theorem).
Note that we do not constrain $\ObvsAlphabet$ to be real-valued---some or all
data may be symbolic. Indeed, in the classic setting for computational mechanics
\cite{shalizi2001computational} $\ObvsAlphabet$ is a discrete and finite
alphabet, typically using only two symbols $\left\{ 0, 1 \right\}$, pasts and
futures are semi-infinite strings and the temporal blocks become words.
Loosening of these constraints is possible even in the theoretical realm
\cite{loomis2023topology}, although this will not be discussed here.

The workhorse of computational mechanics is the predictive equivalence relation.
By predictive equivalence we mean that two specific pasts $\obvspast,
\obvspast'$ correspond to the same distribution over futures $\ObvsFuture$.
This induces an equivalence relation $\EquiFunction{\cdot}$ that maps pasts to
sets of pasts via:
\begin{align}
    \EquiFunction{\obvspast} = \left\{ \obvspast' \in
    \halfSequenceSpaceZero  : 
    \Pr \left( \ObvsFuture \mid \ObvsPast = \obvspast' \right) =
    \Pr \left( \ObvsFuture \mid \ObvsPast = \obvspast \right) \right\} ~.
    \label{eq:equivalenceClass}
\end{align}

If $\obvspast$ and $\obvspast'$ are predictively equivalent, they belong to the
set of pasts induced by their equivalence class. These sets of pasts and their
associated distributions over futures are what we want to find: the elements
of the range of the equivalence mapping are exactly the causal states of the
process. The causal state random variable is denoted $\CausalState$ and its
realization is written $\causalstate$. The set of causal states of a process is
denoted $\CausalStateSet$. (To be more precise, $\CausalStateSet$ is the closure
of the set of causal states, where the closure is taken under convergence in
distribution \cite{loomis2023topology}.)

The cardinality of $\CausalStateSet$ depends on the process under study. On the
one hand---the extreme of utter unpredictability---a sequence of independent
identically-distributed variables trivially is comprised of a single
causal state, because every past $\obvspast$ will induce an identical
distribution over futures. At the opposite extreme of exact predictability, on
the other hand, a discrete period-$k$ process (e.g. $abababa\dots$ has $k = 2$)
has exactly $k$ causal states. In general, the structure of $\CausalStateSet$
is highly variable: the set may be finite or infinite, have countable or
uncountable cardinality, may be well-ordered, or fractal.

The \emph{\eM} of a process is the causal state set $\CausalStateSet$ together
with the transition dynamic over the causal states induced by the shift
operator of the process.
When both $\CausalStateSet$ and $\ObvsAlphabet$ are discrete and finite, the
dynamic over the causal states may be simply expressed with a set of
symbol-labeled transition matrices $\left\{ T^{\left( \obvs
\right)}_{\causalstate, \causalstate'} = \Pr \left( \obvs, \causalstate' \mid
\causalstate \right) : \causalstate, \causalstate' \in \CausalState, \obvs \in
\ObvsAlphabet \right\}$. When $\CausalState$ is finite, then, the \eM
takes on the familiar form of an \emph{edge-emitting finite-state hidden
Markov model}.\footnote{With the additional constraint that the symbol-labeled
transitions are \emph{unifilar}, which is to say that the current state of the process
$\causalstate_t$ and the next observed symbol $\obvs_{t+1}$ uniquely determines
the next state $\causalstate'_{t+1}$.} When $\CausalStateSet$ is infinite but $
\ObvsAlphabet$ is discrete and finite, the transition dynamic may described with
an \emph{iterated function system} \cite{Jurg20b,Jurg20c}. When $ \ObvsAlphabet$
becomes continuous, the process is \emph{hidden semi-Markov} \cite{Marz17b} and the transition dynamic over causal states
becomes a general inhomogeneous stochastic difference equation: the probability
of state-to-state transitions depends both on the state before the
transition and the value $\obvs \in \ObvsAlphabet$ inducing the transition.

By construction, the \eM is the minimal optimally predictive model of a process.
Additionally, \eMs are used to directly calculate in closed form a wide variety
of relevant information theoretic properties, such as the Shannon entropy rate
and the statistical complexity, a measure of the process's complexity
\cite{crut12a}. In addition, being built constructively from conditional
distributions over futures means that \eMs are invariant under bijective
transformations---even nonlinear ones---of the observation space. That is to
say, a physical process measured under a change of reference frame will still
induce the same equivalence classes, making the \eM an intrinsic property of
$\Process$ robust to changes in measurement. These advantages make the \eM
theoretically compelling, but the exactness of the theoretical construction have
made \eMs difficult to work with from an inference perspective.

The recently developed \kcs method \cite{brodu22rkhs} is, to our knowledge, the
first causal state inference algorithm that is practicable for nearly arbitrary
data. It is also well-grounded: proof that causal states are always well defined
and empirical observation-based approximations will converge has been recently
been extended to continuous-valued processes under the mild conditions listed
above \cite{loomis2023topology}. This vastly expands the practicality of
computational mechanics to new classes of processes.

\section{Kernel Hilbert \texorpdfstring{{$\epsilon$}}{e}-machines}
\label{sec:kerneleMachines}

Recent work on the topology and geometry within computational mechanics showed
that the causal states can be naturally embedded into a Hilbert space using an
appropriate kernel \cite{loomis2023topology}. This connects computational
mechanics to the well-studied arena of reproducing kernel Hilbert spaces
(RKHS), setting the stage for the \emph{\keM} (K$\epsilon$M)---an \eM
representation that allows for tractable analysis of more complex processes
than previously possible \cite{brodu22rkhs}.

This section first reviews how causal states naturally form a Hilbert space.
It then recounts the algorithm for \ekcs construction from data. Finally, it
concludes with a walk through the diffusion mapping algorithm, which finds a
low-dimensional embedding of the empirically estimated \kcs set. 

\subsection{Causal states form a Hilbert space}
\label{sec:CausalStateHilbertSpace}

\begin{figure}
    \includegraphics[width=\columnwidth]{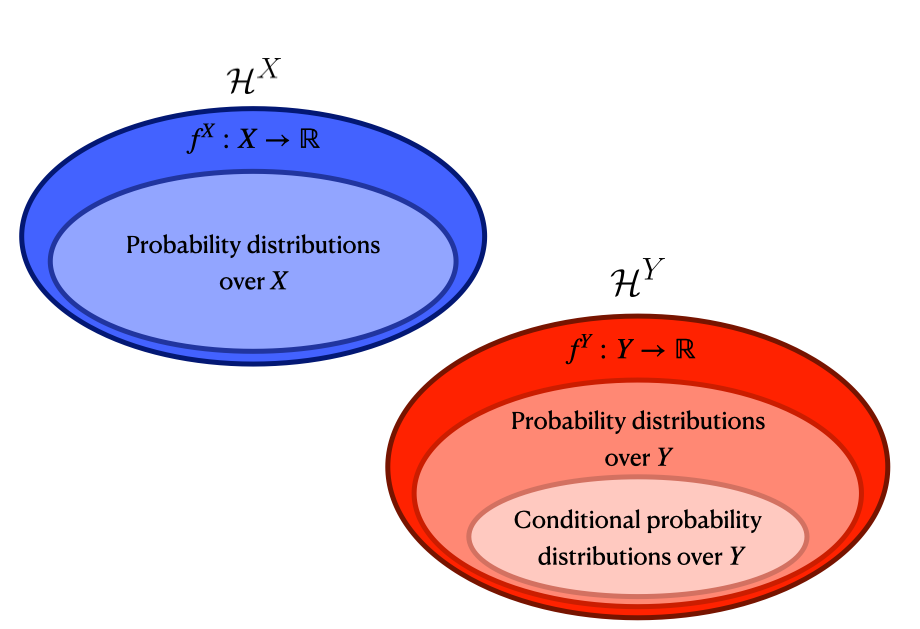}
    \caption{Probability distributions over pasts $\ObvsPast$ and conditional
    distributions over futures $\ObvsFuture$ can be represented as points in the
    reproducing kernel Hilbert spaces $\HilbertSpacePast$ and
    $\HilbertSpaceFuture$, respectively. 
    \label{fig:hilbertSpaces}
    }
\end{figure}

\Cref{sec:computuationalmechanics} defined $\CausalStateSet$ as the
(closure of the) set of the causal states of a process. Now, let
$\CausalStateSpan$ be the vector space of signed measures generated by the
closed span of $\CausalStateSet$. This is the smallest vector space that
contains all the causal states. The dimensionality of $\CausalStateSpan$ may be
much less than the cardinality of the causal state set $\CausalStateSet$. In
general, for any process generated by a hidden Markov model, $\dim
\CausalStateSpan$ will be finite but $\CausalStateSet$ may be uncountably
infinite \cite{loomis2023topology, Jurg20c}. 

The space $\CausalStateSpan$ is a subspace of the space of probability measures
over half-infinite sequences $\ProbSequenceSpace$. By the Riesz representation
theorem, given a symmetric positive-definite kernel $\kernel :
\halfSequenceSpace \times \halfSequenceSpace \to \reals$, we may embed any
measure $\measure \in \ProbSequenceSpace$ into a Hilbert space by writing down
the function:
\begin{align}
    f_{\measure} = \int_{\halfSequenceSpace} \kernel 
    \left( \obvsfuture, \cdot \right)
    d \measure \left( \obvsfuture \right) ~, 
\label{eq:RKHSEmebddingFunction}
\end{align}
which belongs to the reproducing kernel Hilbert space $\HilbertSpace_{\kernel}$.
(The dot notation indicates a free argument so that $\kernel \left( \obvsfuture,
\cdot \right) \in \HilbertSpace_{\kernel} \; \forall \, \obvsfuture \in
\halfSequenceSpace$.) This Hilbert space is generated by the kernel $\kernel$
and equipped with the inner product between measures:
\begin{align}
    \innerProductMath{f_{\measure}}{g_{\measureAlt}}_{\kernel}
    & = \int \int \kernel(\obvsfuture, \obvsfuture') \, 
    d \measure(\obvsfuture) \, d \measureAlt (\obvsfuture') \nonumber \\
     & = \int f_{\measure} (\obvsfuture') \, d \measureAlt (\obvsfuture') ~.
\label{eq:RKHSInnerProduct}
\end{align}

Since $\CausalStateSpan \subseteq \ProbSequenceSpace$, this allows us to embed
the causal states into a reproducing kernel Hilbert space. Let
$\condCausalMeasure{\obvspast}$ be the conditional measure over the set of pasts
$\EquiFunction{\obvspast} \subseteq \halfSequenceSpaceZero$ induced by the
equivalence mapping \cref{eq:equivalenceClass} on an arbitrary past $\obvspast$.
This conditional measure is well-defined and approximations from empirical
observation will converge in distribution to $\condCausalMeasure{\obvspast}$ for
processes meeting the conditions laid out in \cref{sec:computuationalmechanics}
\cite{loomis2023topology,Uppe97a}. Then the \emph{\kcs}
$\kcausalstate{\EquiFunction{\obvspast}}$ is given by:
\begin{align}
    \kcausalstate{\EquiFunction{\obvspast}} = \int_{\halfSequenceSpace} \kernel 
    \left( \obvsfuture, \cdot \right) 
    d \condCausalMeasure{\obvspast} ( \obvsfuture ).
\label{eq:kcausalstate}
\end{align}

When the kernel is \emph{characteristic} this embedding is unique
\cite{Stein02}. When the kernel is \emph{universal}, the kernel-induced inner
product norm metrizes convergence in distribution in the space of measures,
which is to say that $\Vert f_{\measure_n} - f_{\measure} \Vert_{\kernel} \to 0
$ implies $\int f d \measure_n \to \int f \measure$ for $f \in
C\left(\SequenceSpace\right)$ \cite{loomis2023topology,Sriperumbudur2010}. It
has been shown that universal kernels are also characteristic. In practice,
finding a universal kernel typically means working with kernels derived from a
radial basis function, such as Gaussian kernels.

It should be emphasized that the choice of kernel imposes a specific geometry
onto the \kcs set $\kCausalStateSet$. This is, in a sense, the entire point, as
imbuing the causal state set with a geometry allows us to tractably work with
the \eMs of systems with causal state sets too numerous and/or complex to write
down explicitly. However, we wish for this geometry to be in some sense
``natural'' or, at the very least, consistent with both the topology of
convergence in distribution of the causal states and the product topology of
the sequence space $\SequenceSpace$. The choice of kernel and implications for
the imposed geometry will be further discussed shortly and in greater detail in
\cref{app:kernelMetaparameters}.

\begin{figure}
\includegraphics[width=.95\columnwidth]{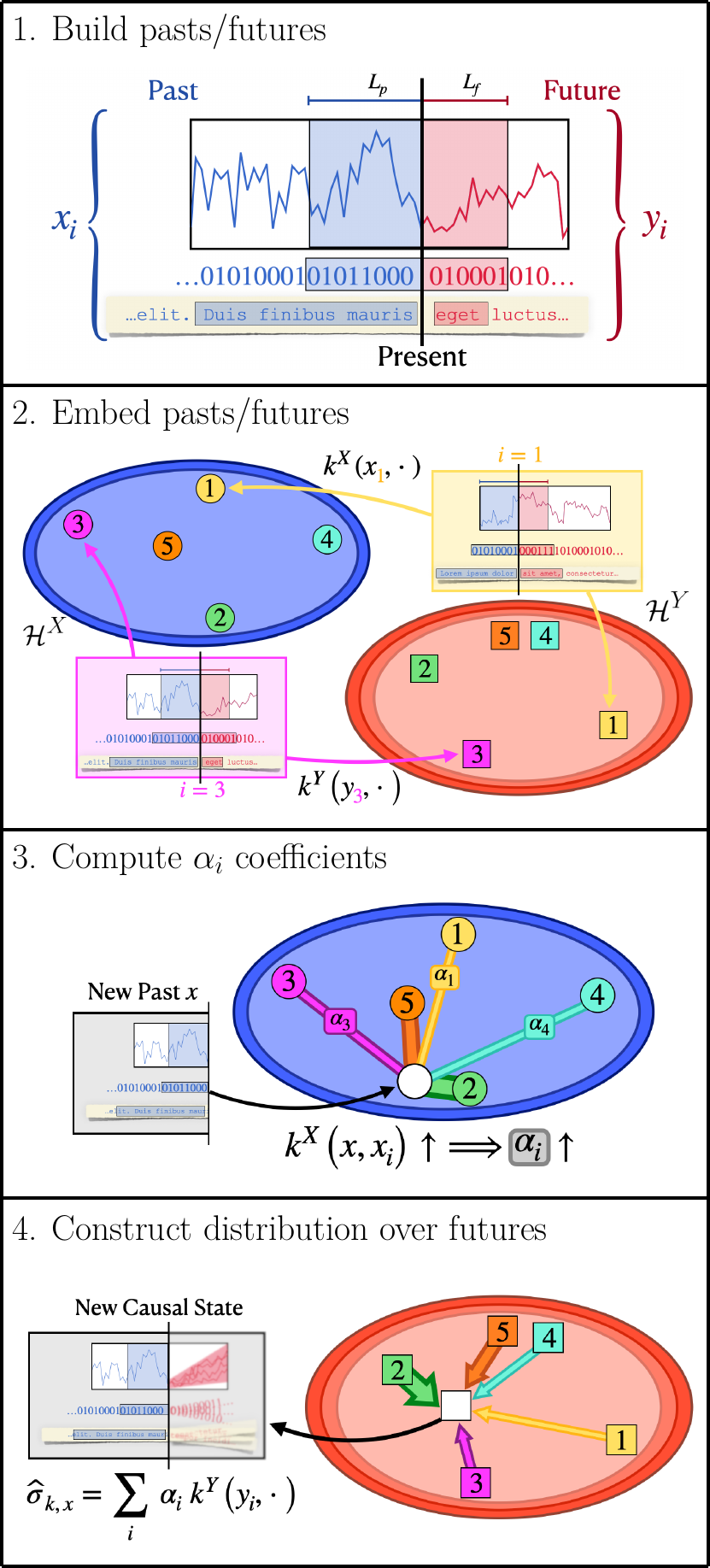}
\caption{The \ekcs algorithm. Step 1: Build a library of past sequences
    $\obvspast_i$ and future sequences $\obvsfuture_i$ from observed data. Step
    2: Construct an appropriate kernel for pasts $\kernelPast$ and a kernel for
    futures $\kernelFuture$. Embed pasts (circles) and futures (squares) into
    their respective Hilbert spaces, $\HilbertSpacePast$ (blue) and
    $\HilbertSpaceFuture$ (red). Step 3: Calculate the \ekcs embedding
    coefficients $\condEmbedCoeff_i$ for a new past $\obvspast$ by comparing
    $\obvspast$ to each embedded past $\obvspast_i$. More similar pasts will
    have larger embedding coefficients. Step 4: Construct the distribution over
    futures for $\obvspast$ using the calculated $\condEmbedCoeff_i$ to weight
    the sum over the associated futures $\obvsfuture_i$. 
\label{fig:algorithmstepthrough}
    }
\end{figure}

\subsection{Empirical \kcs algorithm}
\label{sec:CausalStateEmbeddingAlgo}

In practice, it can be challenging to explicitly build the equivalence classes
$\EquiFunction{\obvspast}$ defined in \cref{eq:equivalenceClass} or to write
down in closed form the conditional measures $\condCausalMeasure{\obvspast}$
for the vast majority of processes. With this motivation, our previous paper
\cite{brodu22rkhs} introduced a method to construct the \ekcss of a process
given a set of observations of total length $\totalLength$. These \ekcss are
guaranteed to converge in distribution to the \kcss in the limit of
$\totalLength \to \infty$, although the nature of this convergence may be
complicated to determine \cite{Smola07, Song13, muandet2017kernel,
loomis2023topology}. 

We construct not just one but two Hilbert spaces---one $\HilbertSpacePast$ over
pasts and the other $\HilbertSpaceFuture$ over futures. This requires
constructing two kernels---$\kernelPast$ and $\kernelFuture$, respectively.
Kernel evaluations are performed on sequences (specifically, pasts or futures).
To conserve the underlying product topology of sequence space, the kernel is
constructed to compare sequences site-wise, with a damping parameter for
indices receding further into the past or future, respectively. That is to say,
when two sequences match indices sufficiently far into the past (future), those
sequences become arbitrarily close under the Hilbert space metric.

This introduces two categories of meta-parameter: first, the specifics of the
site-wise comparison; second, those related to the temporal damping and
aggregation over time. In the first category is the choice of metric (e.g.
Euclidean metric for real-valued data, or the discrete metric for symbolic
data) as well as kernel width or radius. In the second category we consider the
damping method and strength (e.g., exponential or power law) as well as the
method of aggregating time indices (e.g., kernel products or kernel sums). We
will not go into further depth here on the nature of these parameters (see
\cref{app:kernelMetaparameters} for more details) but list them to emphasize
that the construction of the kernel---and thus, imbuing of the causal state set
with a specific geometry---must be done carefully and with consideration to the
underlying system. 

Once the kernel is constructed, though, the empirical estimation of the kernel
causal states is a relatively straightforward four-step algorithm, as depicted
in \cref{fig:algorithmstepthrough}: 
\begin{enumerate}
    \item Assuming that $ \Obvs$ is vector-valued and constructed from
    $\numdatasources$ measurements $\measurementsymbol^i$: $\obvs_t = \left\{
    \measurementsymbol^1_t, \measurementsymbol^2_t, \dots,
    \measurementsymbol^{\numdatasources}_t \right\}$, determine an appropriate
    history length $\historylengthpast$ and future length
    $\historylengthfuture$. The measurements $\measurementsymbol^i$ may include
    heterogeneous data types, see \cref{sec:computuationalmechanics}.
    Pasts and futures are the finite-length temporal blocks
    \begin{align}
        \label{eq:past_seq_def}
        \obvspast_{t} & = \left( \obvs_{t - \historylengthpast +1 },
        \dots, \obvs_{t-1}, \obvs_t \right) \quad \text{and} \\
        \obvsfuture_{t} & = \left( \obvs_{t+1}, \obvs_{t +2}, 
        \dots, \obvs_{t + \historylengthfuture} \right) ~,
        \label{eq:future_seq_def}
    \end{align}
    respectively. We then have $\numObvs$ pasts $\ObvsPastSpace = \left\{
    \obvspast_1, \obvspast_2, \dots, \obvspast_{\numObvs} \right\}$ and futures
    $\ObvsFutureSpace = \left\{ \obvsfuture_1, \obvsfuture_2, \dots,
    \obvsfuture_{\numObvs} \right\}$ where $\numObvs = \totalLength -
    \historylengthpast - \historylengthfuture + 1$.
    
    (Note that we allow $\historylengthpast \neq \historylengthfuture$. See
    \cref{app:kernelMetaparameters} for discussion of measurement-dependent
    history/future lengths.)
    \item Generate two reproducing Hilbert spaces $\HilbertSpacePast$ and
    $\HilbertSpaceFuture$ using two symmetric positive-definite kernels:
    $\kernelPast \left( \obvspast, \cdot \right) : \ObvsPastSpace \to
    \HilbertSpacePast $ and $\kernelFuture \left( \obvsfuture, \cdot \right) :
    \ObvsFutureSpace \to \HilbertSpaceFuture$, respectively. We embed each past
    $\obvspast_i$ into $\HilbertSpacePast$ and future $\obvsfuture_i$ into
    $\HilbertSpaceFuture$ with the appropriate kernel: 
    \begin{align*}
        \widetilde{\obvspast}_{i, k} = & \, \kernelPast ( \obvspast_i, \cdot) ~, \\
        \widetilde{\obvsfuture}_{i, k} = & \, \kernelFuture ( \obvsfuture_i, \cdot)  ~. 
    \end{align*}
    This results in $\numObvs$ many paired points in each Hilbert space. 
    \item Construct the conditional distribution over futures for a specific
    past $\obvspast$ by determining the relative similarity of $\obvspast$ with
    each $\obvspast_i$ and assigning an appropriate coefficient
    $\condEmbedCoeff_i$. One scheme is given by the regularized estimator from
    \cite{Song13}:
    \begin{align}
        \condEmbedCoeff \left( \obvspast \right) = \mathrm{inv}
        \left( \gramMat^\ObvsPast + \regularizationValue \Identity{\numObvs} \right) 
        \kernelvec \left( \obvspast \right) ~,
        \label{eq:conditionalweights}
    \end{align}
    where  $\gramMatpast_{ij} = \kernelPast \left( \obvspast_i , \obvspast_j
    \right)$ is the Gram matrix over past sequences, $\regularizationValue$ is a
    small regularization value, and $\kernelvec \left( \obvspast \right)$ is a
    column vector such that $\kernelvec_i \left( \obvspast \right) = \kernelPast
    \left( \obvspast, \obvspast_i \right)$. More elaborate coefficient schemes
    can be found in \cite{muandet2017kernel}.
    \item Build the \ekcs$\ekcausalstate{\obvspast}$ using the coefficients
    calculated in step 3 \cite[Eq.~11]{Song13}:
    \begin{align}
        \ekcausalstate{\obvspast}
        = \sum_i^N \condEmbedCoeff_i \left( \obvspast \right) 
        \kernelFuture \left(\obvsfuture_i, \cdot \right) ~. 
        \label{eq:embeddingEstimate}
    \end{align}
    The distribution over futures is, in effect, a weighted sum over all
    previously observed future sequence embeddings, where the weight schema
    takes into account the relative similarity of all other observed past
    sequences to the associated past $\obvspast$.
\end{enumerate} 

Steps 3 and 4 are repeated for all past sequences $\obvspast$, resulting
in a set of \ekcss $\ekCausalStateSet = \left\{ \ekcausalstate{\obvspast} :
\obvspast \in \ObvsPastSpace \right\}$. Past sequences that induce the same
distribution over futures will map to the same point in $\HilbertSpaceFuture$ in
the limit of infinite data, reproducing the action of the equivalence class
mapping in \cref{eq:equivalenceClass}. $\ekCausalStateSet$ is not guaranteed to
have any specific geometry beyond lying inside $\HilbertSpaceFuture$.

Algorithm step 3 makes the implicit assumption that similar pasts yield similar
distributions over future. However, we wish to impress upon the reader that
this does not induce a continuity assumption in the sequence space.  For
example, consider two past trajectories $\obvspast$ and $\obvspast'$ on
opposite sides of a boundary between basins of attraction. The embedded points
$\widetilde{\obvspast}_{k}$ and $\widetilde{\obvspast}'_{k}$ may become very
close in $\HilbertSpacePast$, especially with finite-length histories
$\historylengthpast$ and temporal damping. However, if both sides of the basin
are equivalently well-sampled and $\kernelPast$ is adequately discerning, then
the values of each $\condEmbedCoeff_i$ will differ slightly for $\obvspast$ and
$\obvspast'$, with each past preferring its side of the basin. Assuming
$\kernelFuture$ is also sufficiently discerning, the difference in coefficients
is then amplified in step 4 so that the $\obvspast \to
\kcausalstate{\obvspast}$ mapping accurately reflects the discontinuity in the
data space. 

Notice that both $\kernelPast$ and $\kernelFuture$ must be correctly
parametrized to handle discontinuities in the $\obvspast \to
\kcausalstate{\obvspast}$ mapping. This is feasible in principle up to any
chosen resolution provided enough data. An example of a closely related
situation is given in \cref{subsec:Pendulum}. The important point is that no
additional constraint on the structure of the \ekcs set $\ekCausalStateSet$ is
imposed by the algorithm in and of itself. In practice though, the ability to
discriminate fine structures (e.g., fractal boundaries) at any given resolution
depends on both careful parametrization and the availability of sufficient
data.

\subsection{Assigning coordinates to causal states}
\label{sec:diffMapping}

\Cref{sec:CausalStateEmbeddingAlgo}'s algorithm embeds the \kcss using
$\numObvs$ coefficients $\condEmbedCoeff_i$. The causal states, though, are a
property of the process, defined irrespective of the number of observations.
Ideally, this should be reflected in the way their empirical estimates are
represented. In particular, when the process is generated by an ordinary or a
stochastic differential equation, whose phase space is described by
$\numDegreesFreedom$ parameters, then each point in that phase space is its own
causal state \cite{brodu22rkhs}. For these broad classes of processes, the true
causal state set $\CausalStateSet$ thus has infinite cardinality, but it can be
fully indexed by a fixed number $\numDegreesFreedom$ of parameters, independent
of $\numObvs$. An estimate of $\numDegreesFreedom$ can then be recovered from
data \cite{brodu22rkhs}. For generic hidden Markov processes,
$\numDegreesFreedom$ can be arbitrarily high \cite{Marz15a}, although it is in
general finite as noted in \cref{sec:CausalStateHilbertSpace}. When estimating
the causal states from measurements of a physical process, we do not have
access to the microstates and their internal degrees of freedom. Yet, if a
small number of parameters can encode causal states with high accuracy, even
imperfectly, they still hold most of the predictive power at the data scale at
which they are computed. Our previous work \cite{brodu22rkhs} proposed to
encode the \ekcss via a small number of coordinates using a diffusion mapping
\cite{Coifman06}.

This method embeds $\ekCausalStateSet$ into $\reals^{\numDiffCoords}$ using
coordinates constructed from the eigenspectra of a normalized proximity matrix
on the \ekcss. This matrix is often called the \emph{diffusion operator} because
when the underlying data is drawn from a manifold the matrix approximates the
Laplace-Beltrami operator on that manifold. This connection gives an appealing
physical interpretation to the eigenspectra but it should be emphasized that
causal states are not guaranteed or even expected to be drawn from a manifold. 

A major advantage of this method is that we can construct the diffusion operator
using the same kernel over futures used to construct the \kcss. The inner
product on \kcss is given by \cref{eq:RKHSInnerProduct}. To calculate the inner
product on the \ekcss we use the embedding coefficients from
\cref{sec:CausalStateEmbeddingAlgo}: 
\begin{align}
    \innerProductMath{\ekcausalstate{\obvspast}}{\ekcausalstate{\obvspast'}}
    = \sum_l \sum_m \condEmbedCoeff_l ( \obvspast ) \condEmbedCoeff_m (\obvspast')
     \kernelFuture \left( \obvsfuture_l, \obvsfuture_m \right)
    \label{eq:innerproduct_RKHS}
  ~.
\end{align}

With this, we define our \kcs proximity matrix as $\gramMatCS_{ij} =
\innerProductMath{\ekcausalstate{\obvspast_i}}{\ekcausalstate{\obvspast_j}}$. We
then normalize $\gramMatCS$ to eliminate the influence of the sampling density,
resulting in a nonsymmetric row-stochastic matrix $\diffTransMat$. (See
\cite{Coifman06, berry2015nonparametric} for further details.) Intuitively,
$\diffTransMat_{ij}$ gives the probability that a virtual point at
$\ekcausalstate{\obvspast_i}$ would travel to $\ekcausalstate{\obvspast_j}$
under the action of purely isotropic diffusion acting with respect to the
proximity set by the kernel inner product.

We note a potential point of confusion here: This is \emph{not} the true \eM
dynamic over the causal states referenced in
\cref{sec:computuationalmechanics}. Rather, this diffusion is virtual and only
used here as a non-Euclidean measure of the proximity of the points. This
virtual diffusion could, however, be used as a mathematical foundation with
respect to which to quantify properties of the \eM dynamic. Indeed, the
diffusion distance is invariant under permutation of the data indices, hence
can only reflect static properties of $\ekCausalStateSet$.

\begin{figure}
\includegraphics[width=\columnwidth]{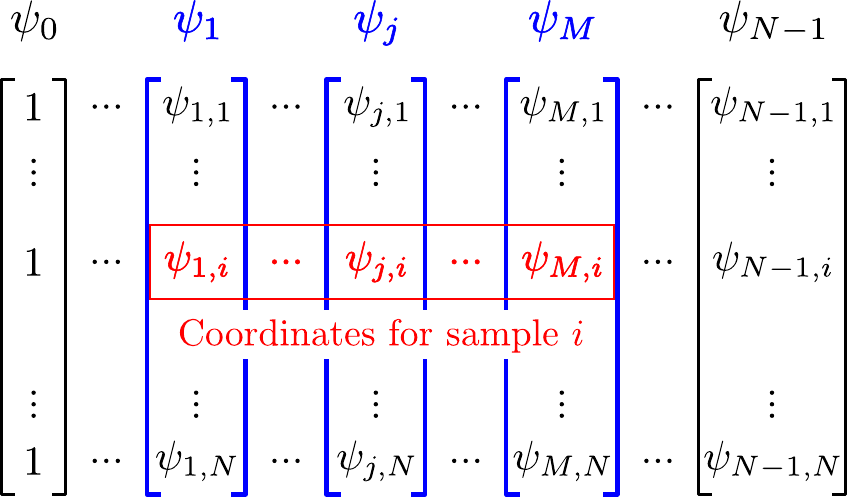}
\caption{Relation between the causal diffusion components $\psi$---the
	eigenvectors of the diffusion matrix $\diffTransMat$---and the coordinates
	that encode the \ekcss.}
\label{fig:eigenvectorsChart}
\end{figure}

Finally, we perform a spectral decomposition on $\diffTransMat$. We call the
right eigenvectors $\diffREigenvector$ the \emph{\cdcs}.

Note that since $\diffTransMat$ is row-stochastic, $\diffEigenvalue_0 = 1$ and
$\diffREigenvector_0$ is constant. We normalize the eigenvectors so that
$\diffcoord{0}{i}=1$ for all $i$, and omit it in the definition of the \cdcs.
The associated left eigenvector $\diffLEigenvector_0$ gives the
diffusion-induced density at each sample
$\ekcausalstate{\obvspast_i}$.\footnote{Note that $1 = \diffLEigenvector_0^T
\diffREigenvector_0 = \sum_i \diffLEigenvector_0$ with this normalization of
$\diffREigenvector_0$, matching the density interpretation.}

We use the coefficients of these eigenvectors as shown in \cref{fig:eigenvectorsChart} to assign coordinates to each $\ekcausalstate{\obvspast_i}$. In this way we embed the \ekcs estimates in $\reals^{\numDiffCoords}$:
\begin{align}
    \ekcausalstate{\obvspast_i}|_\numDiffCoords \equiv \left( \diffcoord{1}{i}, \ldots, \diffcoord{M}{i} \right)
  ~.
\end{align}
The diffusion distance is recovered by scaling these coordinates by the
corresponding eigenvalues of $\diffTransMat$:
\begin{align}
    \diffDist\left( \ekcausalstate{\obvspast_i}|_\numDiffCoords, \ekcausalstate{\obvspast_l}|_\numDiffCoords \right) =
    \sum_{j=1}^{\numDiffCoords} \diffEigenvalue_j^2 \left( \diffcoord{j}{i} - \diffcoord{j}{l} \right) ^2
	~.
    \label{eq:diffDist}
\end{align}

\begin{figure*}
    \includegraphics[width=\textwidth]{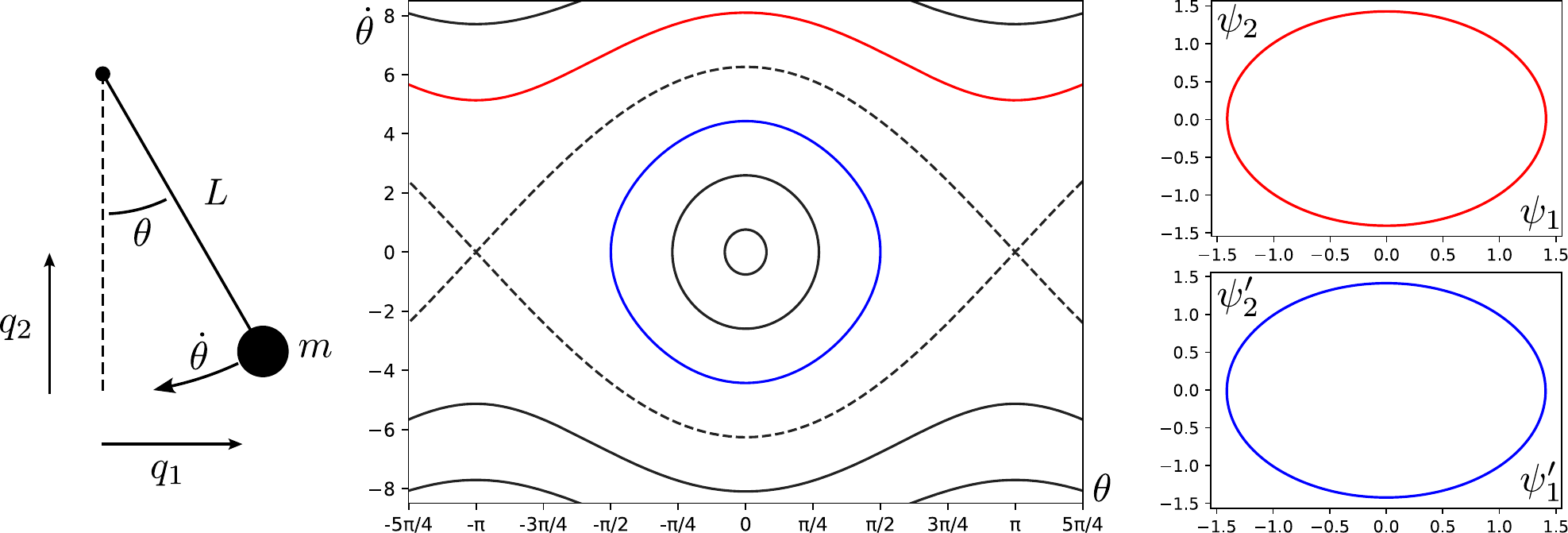}
    \caption{Left: The simple pendulum with angular velocity $\dot{\pendangle}$
    and angle from the vertical $\pendangle$. Middle: Pendulum trajectories in
    the angle and angular velocity $\left( \pendangle, \dot{\pendangle} \right)$
    phase space at different energy levels. The dotted line is the separatrix,
    for which the total energy $E = \equilibriumEnergy$. The ellipsoid
    trajectories in the middle represent trajectories with $E <
    \equilibriumEnergy$ and the trajectories above and below the separatrix have
    $E > \equilibriumEnergy$. The red and blue trajectories were used as input
    to the \ekcs algorithm. Right: Each resultant color coded set of \ekcss
    plotted using their first two diffusion coordinates. Eigenvalues and spectral gaps are given in
    \cref{fig:pendulum_eigenvalues}. }
    \label{fig:undampedPendulum}
\end{figure*}

This distance matches that of the virtual diffusion process corresponding to
$\diffTransMat$; i.e., how ``long'' it takes to ``diffuse'' between two given
\ekcss. This is not the same as the distance in $\HilbertSpaceFuture$. However,
the relation between both can be inferred by expressing the spectral
decomposition of $\diffTransMat$ as that of an operator in
$\HilbertSpaceFuture$. (See also \cite{Coifman06}.)

Note that we retained only $\numDiffCoords < \numObvs$ components. This is
justified for the cases in this section's introduction where $\numDiffCoords$
can be formally inferred. More generally, and for natural processes especially,
we cannot state in advance how many components are needed to encode the causal
states up to a prescribed accuracy. When a spectral gap exists we identify
$\numDiffCoords$ by the eigenvalue decay profile; see examples in our previous
work \cite{brodu22rkhs} and in the next sections. Otherwise, we can set
$\numDiffCoords$ so that the residual distance
$\diffDist\left(\ekcausalstate{\obvspast_i}|_\numDiffCoords,
\ekcausalstate{\obvspast_i}|_{\numObvs-1}\right)$, averaged over all samples,
remains below a given threshold.

Intuitively, each added component captures more predictive information---that
not previously contained in the others. This is similar to how principal
components progressively capture variance in a data set, but for predictive
information instead of variance and using a strongly nonlinear transformation
of the data.  To see this, recall that eigenvalues $1 \geq \diffEigenvalue_j
\geq 0$ are sorted in decreasing magnitude and that \cref{eq:diffDist}
specifies which proportion of the distance between the \ekcss is captured by
each coordinate. However, the \ekcss themselves are embeddings of conditional
distributions over futures. Hence, refining their proximity also means, in some
sense, refining the ability to discriminate between their predictions. This
notion aligns with other embeddings of causal states \cite{Jurg20b,Loom22}, but
with the additional property of ordering the coordinates.

Another interpretation is to recall that the diffusion map is parametrized to
approximate a Laplace operator. In this perspective, the eigenvectors represent
the vibrating modes of a manifold enclosing the \ekcss, with the appropriate
eigenvalues associated to frequencies. Components of the diffusion map with $j >
\numDiffCoords$ are then assimilated to high frequency noise that can be safely
discarded. More generally, the connection between diffusion maps and harmonic
analysis has been well detailed in \cite{coifman2005geometric,
berry2020spectral}. This connection opens up the possibility to compare the
\ekcss of a process estimated at multiple scales, which we keep for future work.

Another point of view on choosing $\numDiffCoords$ is to imagine the signals as
measurements of some unknown physical process we seek to model. It is often not
desirable to embed all observed fluctuations as there may be sensor noise or
additional microscopic phenomena that perturb observations of the underlying
system. In this view, lower frequency components capture the essence of the
dynamics and additional components capture finer and finer details. The goal is
to retain sufficient components to reflect the system's physics, but not those
of irrelevant external factors.

Taking the last three sections together, the \ekcss and the \cdcs define the
structural model class of \emph{kernel \eMs}. The members of this model class
inherit the desirable properties of their analytical brethren---namely
optimality, minimality, and uniqueness. We explicitly left out discussing the
dynamic over the causal states of kernel \eMs for our purposes here, as
our focus is on reporting the successful empirical approximation of the \kcss
$\kCausalStateSet$ for a series of examples using both synthetic and real data.
We leave to the future the task of inferring the \kcs dynamic for the
examples reported here.

\section{Simulated examples}
\label{sec:simulatedexamples}

This section presents two examples of the \ekcs algorithm applied to simulated
data. The first system is the simple pendulum, a low-dimensional deterministic
nonlinear system for which the \eM can be solved exactly. We walk through
finding the causal state set and \kcs set analytically and compare these
results to the \ekcss.

Our second simulated example is data from a high-dimensional molecular dynamics
simulation of the $n$-butane molecule. We show that the \ekcss distinguish
between the three well-known low-energy conformations of the molecule within the
first two \cdcs, with finer differences in molecular positions revealed by
considering higher dimensions in the causal diffusion embedding.

\subsection{The pendulum}
\label{subsec:Pendulum}

For a sufficiently simple system, the causal states and their kernel
equivalents can be written down exactly and compared to the computational
results of the \ekcs algorithm. We demonstrate this using a classical pendulum
as sketched in \cref{fig:undampedPendulum} (left). The pendulum is purely
deterministic and so the structure of the causal state set is especially
straightforward. The purpose here is primarily to introduce notation and build
intuition around working with causal states and kernel causal states for
continuous-valued systems.

\begin{figure*}
    \includegraphics[width=0.32\textwidth]{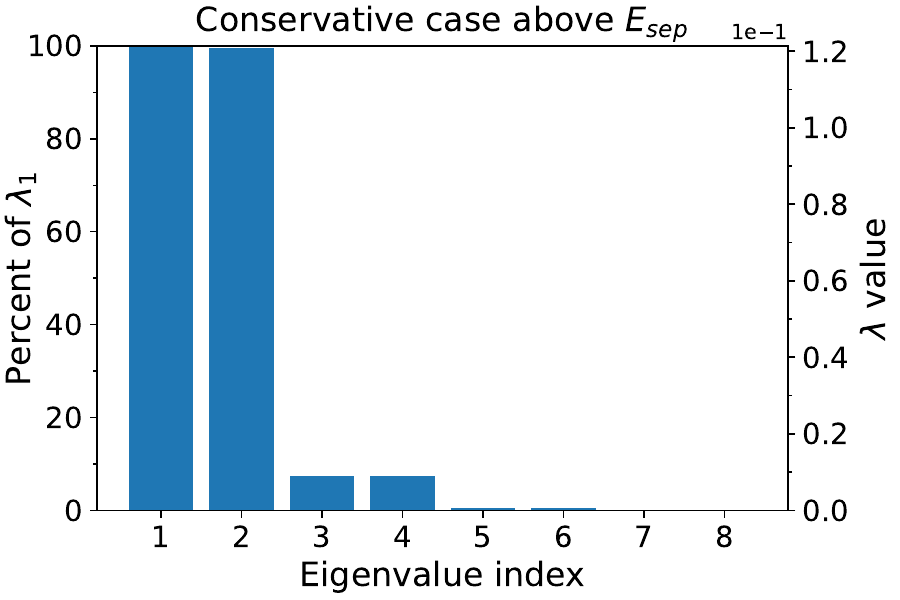}
    \hfill
    \includegraphics[width=0.32\textwidth]{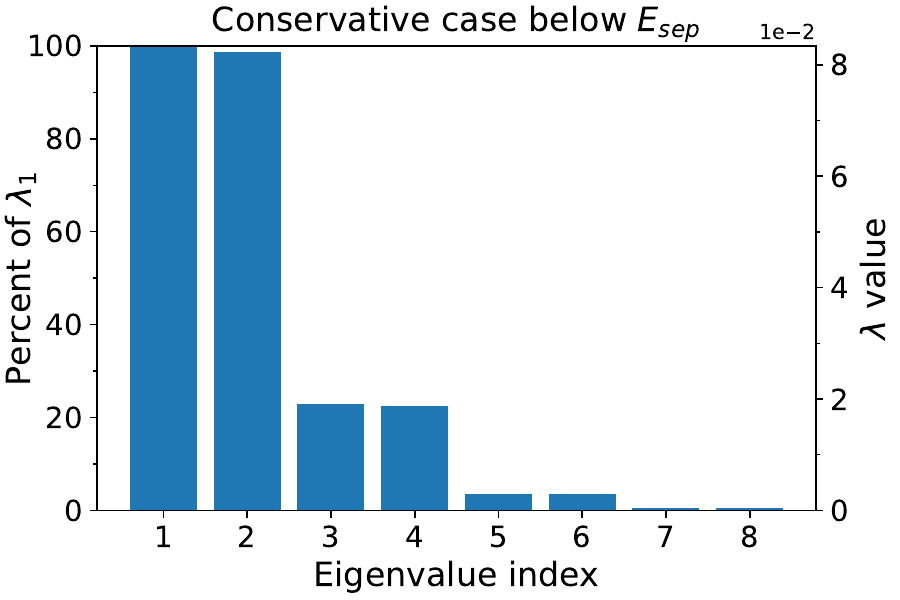}
    \hfill
    \includegraphics[width=0.32\textwidth]{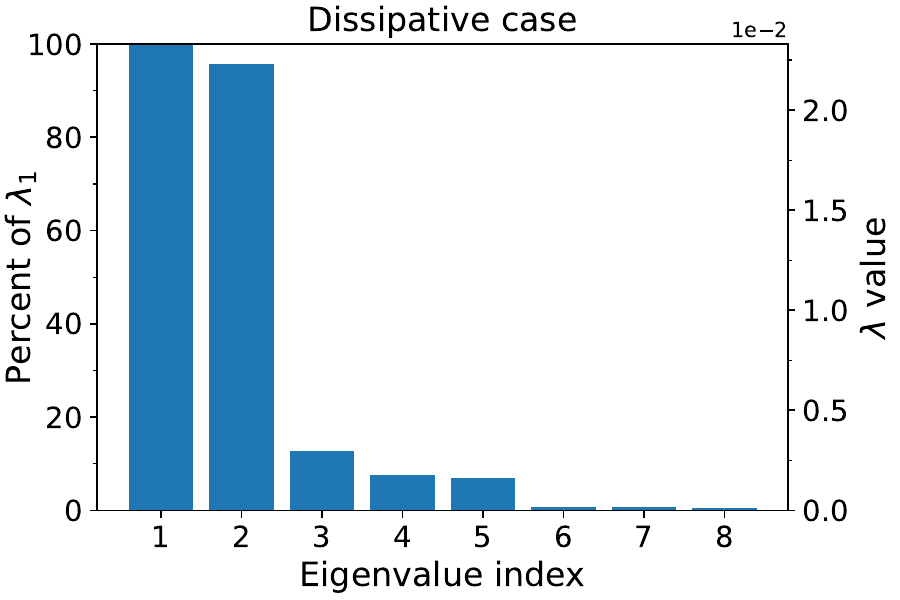}
    \caption{Eigenvalues and spectral gaps for the simple pendulum \ekcss. The two
    conservative cases match those in \cref{fig:undampedPendulum} and the dissipative case \cref{fig:dampedPendulum}.}
    \label{fig:pendulum_eigenvalues}
\end{figure*}

The equation of motion of a pendulum is given by:
\begin{align}
    \frac{ d^{2}\pendangle }{ dt^{2} } = - \frac{g}{L} \sin \pendangle ~, 
    \label{eq:pendulumeqofmotion}
\end{align}
where $L$ is the length of the (massless) pendulum rod, $\pendangle$ is the
angle of the pendulum with the vertical, and $g$ is strength of gravity. We
obtain a set of coupled first-order equations by considering the angular
velocity $\dot{\pendangle}$ as a separate variable:
\begin{align*}
    \frac{d\pendangle}{dt}       & = \dot{\pendangle} \\
    \frac{d\dot{\pendangle}}{dt} & = - \frac{g}{L}\sin\pendangle
  ~.
\end{align*}
The position of the pendulum at time $t$ relative to an origin placed at the
bottom of the swing is given by the generalized coordinates:
\begin{align*}
    \pendcoord_1 (t) & = L \sin \pendangle (t) \\  
    \pendcoord_2 (t) & = L \left( 1-  \cos \pendangle (t) \right) ~. 
\end{align*}

Despite its simplicity, the system of differential equations cannot be solved
in terms of elementary functions without the small angle approximation, which
reduces the dynamics of the pendulum to simple harmonic motion. However, the
solutions can be computed numerically. A phase portrait of pendulum
trajectories in $\left( \pendangle, \dot{\pendangle} \right)$ at different
total energy levels is given in \cref{fig:undampedPendulum} (middle).

There are two modes of behavior: First, a pendulum without sufficient energy to
rotate entirely around the pivot, in which case the phase portrait is a limit
cycle that grows increasingly elliptical as energy is increased; second, a
pendulum with enough energy to swing all the way around, in which case the
phase portrait is a sinusoidal curve. These modes are separated by a boundary,
called a \emph{separatrix}, found at the energy level $\equilibriumEnergy$ at
which the pendulum can stand straight up in an unstable equilibrium at
$\pendangle = \ang{180}$. 

Now, let's construct the causal state set of an idealized pendulum that has
been swinging for infinite time at energy level $\totalEnergy \neq
\equilibriumEnergy$. Since there is no stochasticity in the system, this is
simple: each point on the phase space trajectory corresponds to exactly one
causal state. Let us parameterize the phase space trajectory with a single
coordinate denoted $\pendCSParam \in \left[0 , 2 \pi \right)$. When
$\totalEnergy \leq \equilibriumEnergy$ we use the angle from the positive
horizontal axis for this purpose; when $\totalEnergy > \equilibriumEnergy$ we
use $\pendangle \mod 2 \pi$. Either way, each causal state
$\causalstate_{\pendCSParam}$ is labeled by its value of $\pendCSParam$.

The time indices for which the pendulum will be at exactly $\pendCSParam$ are
given by $\tau_{\pendCSParam} = \left\{ \uptau + n \period : n \in \integers
\right\}$, where $\uptau$ is the time it takes the pendulum to travel from its
initial condition to $\pendCSParam$ and $\period$ is the period of the
oscillation. For our idealized pendulum, each causal state
$\causalstate_{\pendCSParam}$ is associated with a set of infinitely many pasts
$\left\{\obvspast_{\tau} : \tau \in \tau_{\pendCSParam} \right\}$. Since the
pendulum is deterministic, each past in this set induces not only the same
\emph{distribution} over futures $\condCausalMeasure{\pendCSParam}$ as required
by the equivalence relation but also the exact same \emph{realization} of the
future: the exactly solvable trajectory of the pendulum for $t > \tau, \, \tau
\in \tau_{\pendCSParam}$. 

So the causal state set $\CausalStateSet$ of the simple pendulum has infinite
cardinality but is one dimensional, in the sense that we can uniquely label each
casual state with a single variable. It should be noted that this analysis does
not depend on the total energy of the pendulum.\footnote{Except in the case of
the unstable equilibrium. For a pendulum that stands at $\pendangle = \ang{180}$
indefinitely, the recurrent causal state set is a single point.}

Now let's construct the \kcss. Recalling that the conditional distributions
over futures $\measure_{\pendCSParam}$ are delta distributions over a single
future, we have: 
\begin{align*}
    \causalstate_{k, \pendCSParam} = 
    \int_{\halfSequenceSpace} \kernelFuture \left( \obvsfuture, \cdot \right) 
    d \measure_{\pendCSParam} ( \obvsfuture ) = 
    \kernelFuture \left( \obvsfuture_{\pendCSParam}, \cdot \right) ~. 
\end{align*}

Following \cref{sec:CausalStateHilbertSpace}, the inner product between the
\kcss of the pendulum reduces to kernel evaluations between futures: $\left<
\causalstate_{k, \pendCSParam} , \causalstate_{k, \pendCSParam'} \right> =
\kernelFuture \left( \obvsfuture_{\pendCSParam}, \obvsfuture_{\pendCSParam'}
\right)$. If we assume that the kernel is constructed to agree with the
Euclidean distance between positions of the pendulum in physical space, then
kernel causal states corresponding to nearby points on the pendulum trajectory
in phase space will be close to each other in the kernel Hilbert space.
Furthermore, the \kcs set $\kCausalStateSet$ will inherit the topology of the
trajectory in phase space, which is to say the circle $S^1$.

\begin{figure*}
\includegraphics[width=\textwidth]{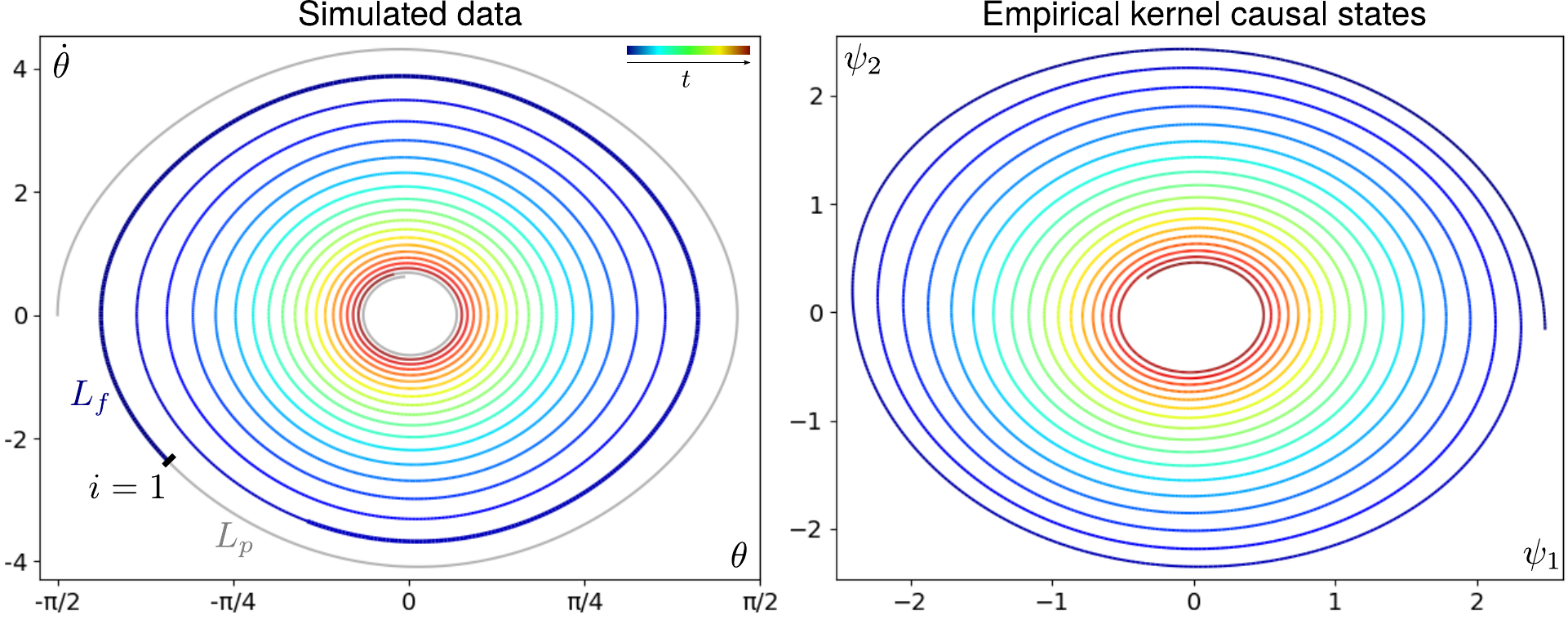}
\caption{Left: The pendulum trajectories in the angle and angular velocity
    $\left( \pendangle, \dot{\pendangle} \right)$ parameter plane colored by
    time. The gray sections at the beginning and end are the first
    $\historylengthpast$-length past and last $\historylengthfuture$-length
    future, respectively, and so do not correspond to any embedded causal
    states. The bold segment denotes the $\historylengthfuture$-length future
    for the first embedded history. Right: The \ekcss plotted using their
    first two diffusion coordinates.
}
\label{fig:dampedPendulum}
\end{figure*}

Let's compare this to the structure of the \ekcs set $\ekCausalStateSet$.
Consider the \ekcss of the pendulum at two different energy levels, as plotted
in \cref{fig:undampedPendulum} (middle). The first has $E < \equilibriumEnergy$
and is given in blue, the second is given in red and has $E >
\equilibriumEnergy$. In both cases we use the generalized coordinates $\left(
\pendcoord_1, \pendcoord_2 \right)$ as input to the \ekcs algorithm and take
$\historylengthfuture = \historylengthpast = \period$ so that our past and
future sequences are arrays of shape $\period \times 2$:
\begin{align*}
    \obvspast_t  &= \left( 
        \begin{bmatrix} \pendcoord_{1, t - \period +1} \\ 
        \pendcoord_{2, t - \period +1} \end{bmatrix} , 
        \dots, 
        \begin{bmatrix} \pendcoord_{1, t -1} \\ 
        \pendcoord_{2, t -1} \end{bmatrix} ,
        \begin{bmatrix} \pendcoord_{1, t } \\ 
        \pendcoord_{2, t 1} \end{bmatrix} 
    \right)  \\ 
    \obvsfuture_t &= \left( 
        \begin{bmatrix} \pendcoord_{1, t +1} \\ 
        \pendcoord_{2, t +1} \end{bmatrix} , 
        \begin{bmatrix} \pendcoord_{1, t +2} \\ 
        \pendcoord_{2, t + 2} \end{bmatrix} ,
        \dots ,
        \begin{bmatrix} \pendcoord_{1, t + \period } \\ 
        \pendcoord_{2, t + \period} \end{bmatrix} 
    \right)  ~. 
\end{align*}
We use a Gaussian kernel with variance \num{1} for comparing values in each
sequence and a product of such kernels for both $\kernelPast$ and
$\kernelFuture$.

After the \ekcss are constructed, they are each embedded using the diffusion
transform from \cref{sec:diffMapping}. The two cases are plotted using their
respective first two components in \cref{fig:undampedPendulum} (right) and color
coded to match their respective trajectories in the phase space plot. Although
the \ekcs sets are point sets, in the plot the $\ekCausalStateSet$ sets appear
as smooth ellipsoids. This is due to the density of the sampling of the pendulum
trajectories. If this sampling was reduced, we would begin to observe gaps
between the \ekcss, as we do in the real data examples in
\cref{sec:realdatapps}.

As is visually apparent, the \ekcss $\ekcausalstate{\obvspast}$ and the
diffusion eigenvectors are identical for both $E > \equilibriumEnergy$ and $E<
\equilibriumEnergy$. This is just as expected---as noted above, in nearly every
case the causal states are not dependent on the pendulum's energy level, and so
the \ekcs do not distinguish between these two cases.

In both cases we require two components $\left(\diffREigenvector_1,
\diffREigenvector_2 \right)$ to embed the \ekcss. This is because the diffusion
mapping algorithm retains the topology of the \kcss when embedding them in
Euclidean space and $S^1$ is minimally embedded in $\reals^2$. We indeed note a
spectral gap with the third eigenvalue in \cref{fig:pendulum_eigenvalues}.

The difference between the two cases is explained by the difference in the
nature of the trajectories in $(\pendcoord_1, \pendcoord_2)$ space. Below the
separatrix, the points near each extrema of the pendulum, with $\pendcoord_1<0$
or $\pendcoord_1>0$ but with the same $\pendcoord_2$, are also the points where
the pendulum has the lowest velocity. Hence the trajectories, sampled at
constant rate, consist of more points near these top positions than at the
bottom. The kernel bandwidth thus has to be reduced in order to correctly
separate these points, which become closer and closer as we approach the
separatrix. Above the separatrix, this is not so much of an issue, because the
pendulum never swings back. So, past-future sequences are similar only to those
at the same location along the full loop.

We also wish to consider the case of the damped simple pendulum, in which total
energy of the system changes over time. The equation of motion is adjusted to
add a drag term:
\begin{align}
    \frac{ d^{2}\pendangle }{ dt^{2} } = 
    -\frac{b}{m} \frac{d \pendangle}{dt} - \frac{g}{L} \sin \pendangle ~, 
    \label{eq:dampedpendulumeqofmotion}
\end{align}
where $m=1$ is the mass of the pendulum bob and $b=0.1$ is the drag
coefficient. The trajectory of a pendulum bob originating at $\pendangle_0 =
-\frac{\pi}{2}$ and zero initial velocity is shown in $\left( \pendangle,
\dot{\pendangle} \right)$ phase space in \cref{fig:dampedPendulum} (middle).

\begin{figure*}
\includegraphics[width=\textwidth]{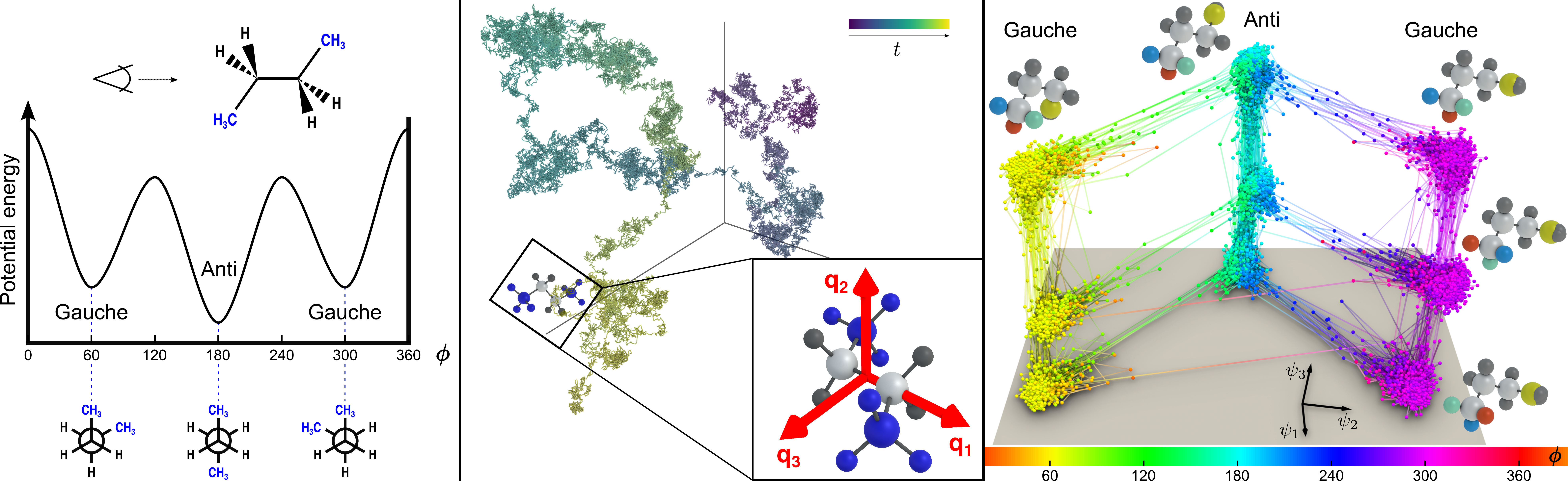}
\caption{Left: The potential energy of $n$-butane molecule plotted along the
	dihedral angle $\dihedralangle$. The Newman projection for each of the
	three low energy conformations is shown below the plot. Middle: The global
	trajectory of the simulated $n$-butane molecule under Brownian motion. The
	inset shows the local frame of the molecule position used as input to the
	\ekcs algorithm. Only the carbon in the back is mobile in this local frame.
	Right: The \ekcs set of the $n$-butane molecule, embedded using the first
	three \cdcs. The frame shows the base vectors in $\reals^3$, scaled down to
	fit in the figure. These give the directions along which the values
	$\diffcoord{j}{i}$ are plotted, for $j=1,2,3$ indicated on the figure and
	$i$ each data index. Dots are plotted at the locations of each
	$\ekcausalstate{\obvspast_i}$ and lines connect successive samples in time.
	Nine clusters are clearly apparent. The average position of each atom in a
	selection of these clusters is shown, with the free carbon (yellow) and
	hydrogens of the fixed methyl group (red, green, and blue) highlighted.
	}
\label{fig:n-butane-molecule}
\end{figure*}

As in the undamped case, each point in the phase space trajectory corresponds
to exactly one causal state. However, in this case, each point is only visited
once and corresponds to exactly one time index---except for the final point at
the bottom of the oscillation, which the pendulum reaches as $t \to \infty$ and
then remains at indefinitely.\footnote{As in the $\equilibriumEnergy$ case, the
recurrent causal state set is a single point.} We call these \emph{transient
causal states}. However, the advantage of the \ekcss algorithm is that because
it does not rely on clustering or frequentist estimation of probability
measures, it is just as capable of constructing the transient causal states, as
we show here. 

As is visually clear, the damped pendulum is quasi-periodic, with the duration
of each swing decreasing over time due to drag. We take the length of the past
and future to be the average quasi-period $\periodAvg$ as inferred over the
full simulation, so $\periodAvg = \historylengthpast = \historylengthfuture$.

The right side of \cref{fig:dampedPendulum} shows the \ekcss of the pendulum
plotted using their first two components $\left(\diffREigenvector_{1},
\diffREigenvector_{2} \right)$. Once again, only two components are necessary
to embed the pendulum's $\ekCausalStateSet$ (\cref{fig:pendulum_eigenvalues}).
In this case we more clearly observe a mild distortion in the \ekcss as
compared to the trajectory in phase space. This is expected, as the causal
states are independent of the frame of reference of the underlying system and
so the diffusion map algorithm (or indeed, any dimension reduction algorithm)
applied to the \ekcss will not return the original system's coordinate system
in general.

Here we do not discuss the dynamics of the \eM or its kernel variant,
intentionally, for now, limiting our scope, which focuses only on the structure
of the \kcs set. However, in this case it is unsurprisingly rather
straightforward: the causal states smoothly evolve along the causal state
parameter $\pendCSParam$. The same holds true for the \kcss. For the \ekcss, we
could in principle infer the evolution of $\ekcausalstate{\obvspast}$ in terms
of the diffusion coordinates using a variety of algorithms
\cite{crutchfield1987equations, baddoo2022kernel, brunton2016discovering,
gauthier2021ngrc, mangiarotti2019gpom}. However, we leave this to the future.

\subsection{n-Butane molecule}
\label{subsec:Molecular-dynamics}

Now, let's move on to a discovery challenge that showcases how the \ekcs method
applies to a high-dimensional, stochastic system. The $n$-butane molecule
$\mathrm{H}_3 \mathrm{C} - \mathrm{CH}_2 - \mathrm{CH}_2 - \mathrm{CH}_3$
consists of four carbon atoms and ten hydrogens arranged as depicted in
\cref{fig:n-butane-molecule}. The energy landscape and conformations of
$n$-butane are well known and characterized by the dihedral angle
$\dihedralangle$ between the two methyl groups. There are three low-energy
conformations: the \emph{antiperiplanar} conformation, for which
$\dihedralangle = \ang{180}$ and the methyl groups are anti-aligned, and the
two \emph{gauche} conformations (from left to right, $\dihedralangle =
\ang{60}$ and $\dihedralangle = \ang{-60}$) for which the methyl groups are
staggered. As we will show, the \ekcss find these conformations even without
taking the dihedral angle as input to the algorithm.

Molecular dynamics simulators begin with the initial positions and velocities of
a set of atoms and simulate their motion over time according to the influence of
atomic interaction forces and possibly external forces (an applied temperature
gradient or magnetic field, for example). The data set we use is the full time
series of the positions of the fourteen atoms of the $n$-butane molecule as
calculated by the molecular dynamics simulator AmberTools15. The $n$-butane
molecule was simulated for an interval of \qty{10}{ns} with a step-size of
\qty{2}{fs}. The resulting data was then downsampled by a factor of \num{100},
of which we retain a trajectory of \num{25000} data points. See
\cite{klus2016operators} for details, where this data set was first
introduced.

The molecule's trajectory in physical space is plotted in
\cref{fig:n-butane-molecule} (middle). The Brownian motion of a molecule at a
constant nonzero temperature is typically modeled with the addition of a
stochastic term---i.e., the Wiener process---rather than explicitly calculating
the collisions with surrounding molecules. As such, this dataset is
intrinsically stochastic, in contrast to the previous examples of purely
deterministic systems. However, the presence of Brownian motion does pose one
complication. If we analyze the movement of the molecule using the fixed
simulation frame of reference, we will model not only the relative atomic
positions---which determines the conformation---but also the global position of
the molecule, which is irrelevant to the conformational dynamics.

To factor this this out, we introduce a local frame of reference internal to
the molecule. The middle carbon bond is used to define one unit vector, see
\cref{fig:n-butane-molecule} (middle) for reference. The second basis vector is
formed by the cross-product of the first with another carbon bond vector, and
the third is computed as the cross-product of the first two vectors. All atom
positions are expressed in that local basis, rendering our past and future
sequences independent from the global location and orientation of the molecule
in space. Thus, the inputs to the causal-state embedding algorithm are
sequences consisting of \num{42} dimensional (three spatial coordinates times
fourteen atoms) position data over one time step, for both the past and future
sequences. 

\begin{figure*}
    \includegraphics[width=0.32\textwidth]{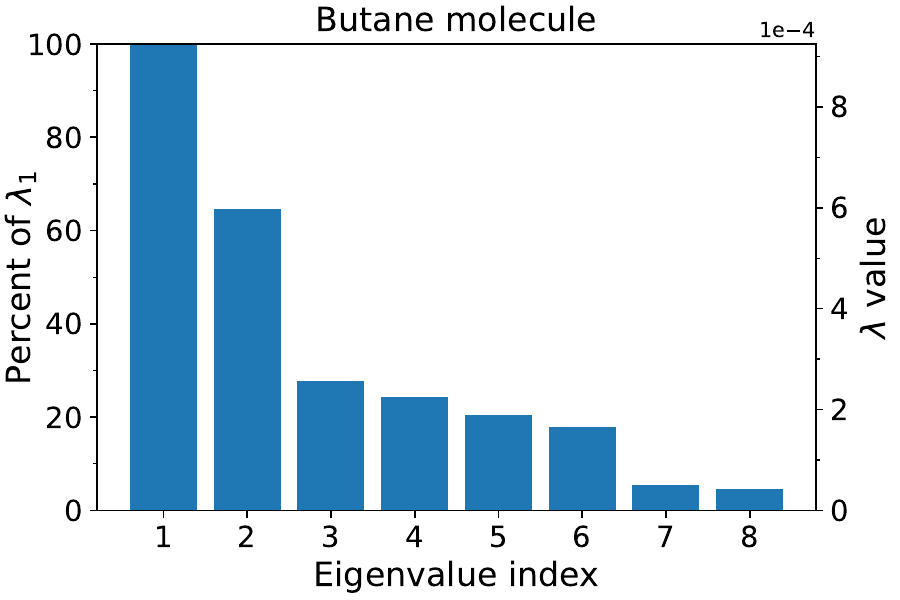}
    \hfill
    \includegraphics[width=0.32\textwidth]{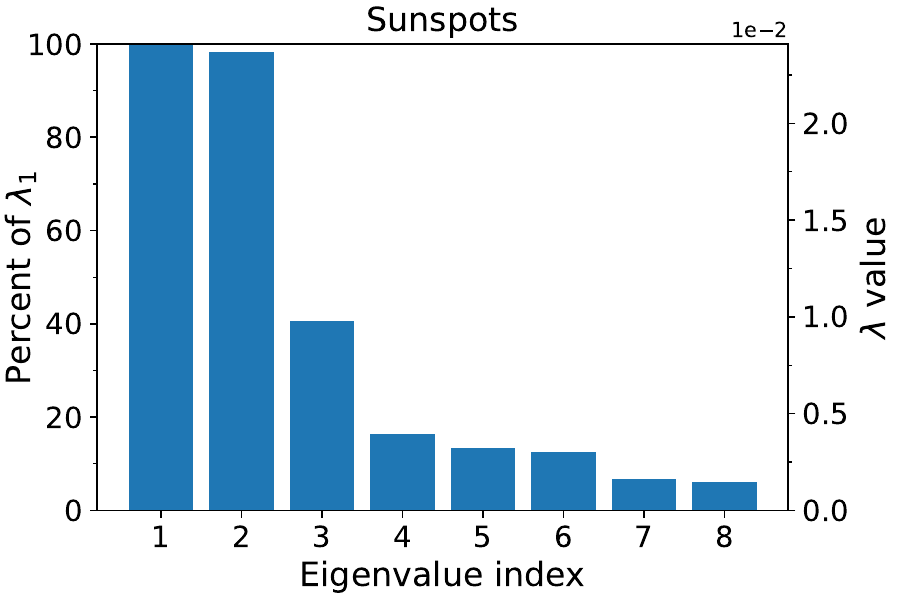}
    \hfill
    \includegraphics[width=0.32\textwidth]{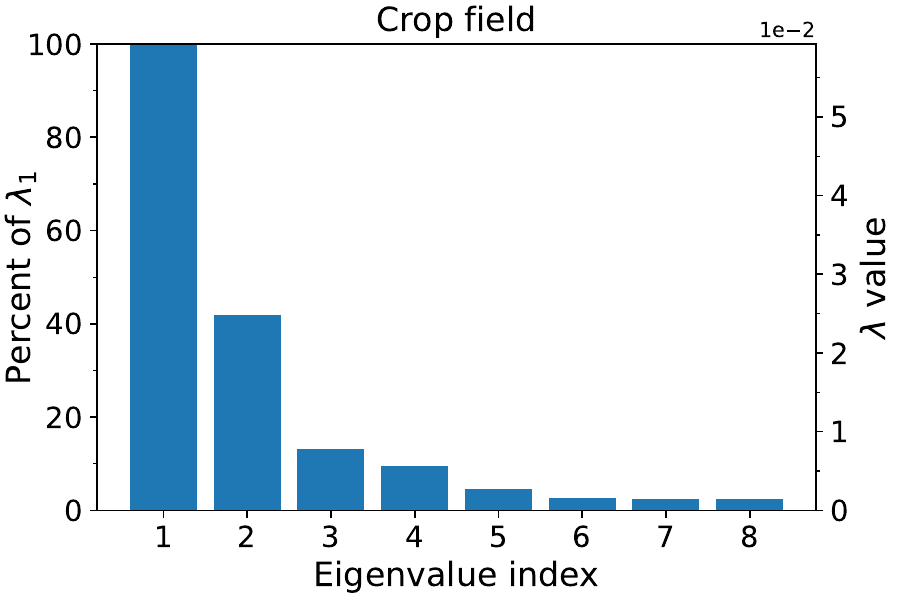}
    \caption{Eigenvalues and spectral gaps for the \ekcss of the butane molecule
    (\cref{fig:n-butane-molecule}), the sunspots (\cref{fig:Sunspots-data}) and
    the crop example (\cref{fig:crop_results}).}
    \label{fig:examples_eigenvalues}
\end{figure*}

The \ekcss are shown in \cref{fig:n-butane-molecule} (right) using their first
three diffusion coordinates. They are colored according to the the dihedral
angle of the molecule during the past associated with that \ekcs (since the
history length is one time step, this is a unique value). The points in the
figure are the \ekcs locations and the lines connect
$\ekcausalstate{\obvspast_t}$ to $\ekcausalstate{\obvspast_{t+1}}$ for all $t$.

The immediately obvious structure of the $\ekCausalStateSet$ is nine clearly
identifiable conformational clusters. Recall that in the limit of identical
predictions over futures the \kcss map to the exact same point in Hilbert space
and that the diffusion mapping translates the difference in the predictions
over futures into distance in the \coords space. This means that a tight
cluster in the \coords space represents a group of time indices in the observed
trajectory where the predicted future was very similar. Jumping between these
clusters then represents a significant change in the next predicted molecular
conformation. Typically, we can see that these transitions occur over one or
two time steps.

The nine \ekcs clusters are arranged into three super-clusters along
$\diffREigenvector_2$ of three sub-clusters along $\diffREigenvector_3$. We
depict the position of each atom for five of these clusters, averaged over
their matching time indices. (The other four clusters follow a similar pattern
and can be reproduced with the provided code.) First consider only the position
of the freely moving carbon as dictated by the local reference frame. This is
the carbon highlighted in yellow in the ball-and-stick depictions. Comparing
its position across the super-clusters shows $\diffREigenvector_1$ separates
the gauche and antiperiplanar conformations, as is also apparent using the
coloring scheme. It does not distinguish between the two gauche
conformations---recall that both gauche conformations have the same potential
energy, as shown in \cref{fig:n-butane-molecule} (left). If we take into
account the second component $\diffREigenvector_2$, the gauche conformations
are separated. And, this gives us an interpretation of the superclusters: all
three low-energy conformations are uniquely associated with a supercluster in
the \ekcs set. Together, the first two components describe the position of the
carbon chain.

Now, consider the positions of the hydrogens attached to the fixed carbon
(highlighted in \cref{fig:n-butane-molecule} (right) in red, green, and blue).
These positions do not change when comparing across the top subcluster of each
supercluster. Only by comparing the subclusters of a single supercluster does
it become clear: the subdivisions in the conformational superclusters along
$\diffREigenvector_1$ represent changes in the orientation of the fixed methyl
group. The first three components shown in \cref{fig:n-butane-molecule} do not
capture the second methyl group orientation. This is why the positions of the
hydrogen atoms of that methyl group are not distinguished in the stick-and-ball
plots. Further components (\cref{fig:examples_eigenvalues}, left) add
subclusters that do distinguish the orientation of the second methyl group.
This preference for tracking the orientation the fixed carbon methyl group is
purely algorithmic: the orientation of the methyl group attached to the freely
moving carbon atom is washed out, in a sense, by the more relevant motion of
the carbon. This comes down to the way one defines the local frame and does
not represent a real asymmetry in the chemical importance of the two methyl
groups.

We note that the frame in \cref{fig:n-butane-molecule} (right) could be slightly
rotated to better align $\diffREigenvector_2$ with the dihedral angle. This 
would also give the gauche and antiperiplanar subclusters the same vertical
coordinates. Recall, however, that the \ekcs algorithm has no notion of potential energy, dihedral angle, or methyl group.

In particular, $\diffREigenvector_3$ separates out the rotation of fixed carbon
methyl group not because there is a \emph{physical} meaning to its orientation,
but because its orientation probabilistically influences the future of the
entire molecule. As is clear in \cref{fig:n-butane-molecule} (right), it is
possible for the molecule to transition between methyl group orientations, and
it is possible for the molecule to transition between conformations, but it is
highly improbable for these transitions to happen in the same time step. The
takeaway is that while structural analysis of the $\ekCausalStateSet$ can be
very powerful, the interpretation of the uncovered structures and any possible
physical meaning of the \cdcs must be done in conjunction with discipline
knowledge.

Comparing this to the simple pendulum in \cref{subsec:Pendulum} demonstrates
the our method's ability to handle highly heterogeneous types of systems. The
structure of the $n$-butane molecule \ekcs set is totally noncyclical and
highly stochastic.

Indeed, the \ekcss naturally organize into something that looks very much like
a finite-state machine. Modeling this process at a larger time scale, we could
discretize each cluster as distinct ``meta''-causal states, with their own
probability distribution of dwell times and instantaneous transitions to the
next cluster. This would give a continuous-time, discrete-state \eM\ of a
renewal process; see \cite{marzen2015informational, marzen2017renewal}. Though
out of present scope, it will be considered in future explorations of nested
causal structure, together with a comparison of the dwell times in each
conformation with the theoretical distribution for residency in each energy
well.



\section{Real data applications}
\label{sec:realdatapps}

This section presents two examples of the kernel causal-state embedding
algorithm applied to real data. The first is a classic and well-known example
in observational astronomy: the number of sunspots visible on the sun's surface
each month. We show that the \cdcs discover several known empirical patterns in
the sunspot sequence. The second example is data from a crop field. This
example shows the use of multi-variate and heterogeneous measurements, some of
which exhibit missing values and low data quality. We show that the \ekcss
detects changes in crop species over a decade-long observation period.

\subsection{Sunspots}
\label{subsec:Sunspots}

\begin{figure*}
    \includegraphics[width=\textwidth]{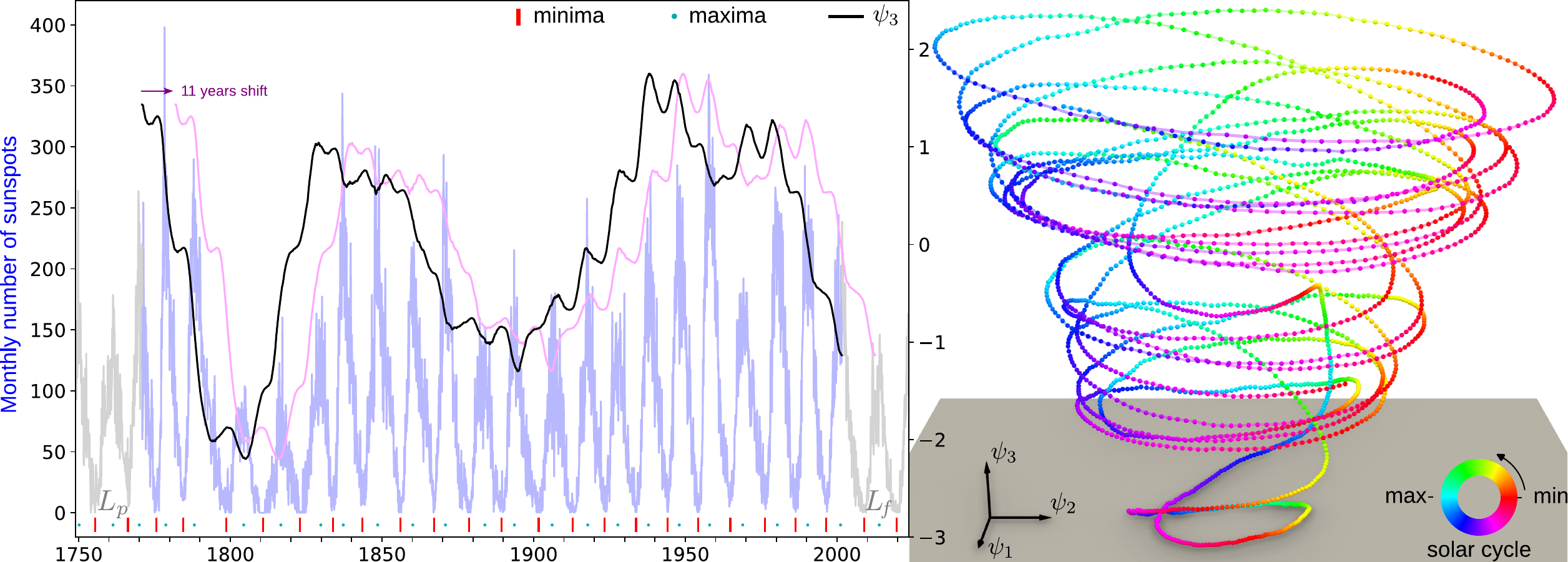}
    \caption{Left: Monthly number of sunspots since 1749. Dates of solar minima
        and maxima are indicated below the main plot. The third diffusion component
        is overlaid on the data in black. Shifting this component by eleven years
        clarifies that $\diffREigenvector_3$ approximately predicts the amplitude
        of the next solar cycle. Right: The \ekcss of the sunspot sequence,
        embedded using the first \cdcs. The indicated frame, dots and lines in the
        figure have the same interpretation as in \cref{fig:n-butane-molecule}. The
        color map indicates the solar cycle phase at the time $t$ associated with
        each $\ekcausalstate{\obvspast_t}$. This phase was computed by linearly
        interpolating the time between each solar minima and maxima.
        }
    \label{fig:Sunspots-data}
    \end{figure*}

Let's start with a deceptively simple and well-known example. One of the
longest-running time series of direct observations without interruptions
available is the number of sunspots. Sunspots have been observed on the sun's
surface since the advent of the telescope. Sunspots are temporary dark spots on
the Sun's surface caused by concentrations of magnetic flux. High-quality
monthly counts are provided by the Sunspot Index and Long-term Solar
Observations (SILSO) databank maintained by the Royal Observatory of Belgium
\cite{sunspotsref} from 1749 to the present day, as shown in
\cref{fig:Sunspots-data} (left). Regularizations are performed to account for
differences in measurement technique over time. 

Driven by the Sun's turbulent convection, the sunspot sequence is a well-known
benchmark in nonlinear time series analysis and it is known to be very
difficult to predict \cite{dang2022comparative}. Fundamental aspects of how and
why the number of sunspots vary over time---and how those changes are related
to solar physics---are not well-understood to this day. Our goal is not to
build a full predictive model, but rather to show how the geometry of the
embedded kernel causal states aids in a structural analysis of the underlying
dynamical system. 

Sunspots are one experimental observation of the broader solar magnetic
activity cycle. This cycle is also reflected in the quasi-periodic variation in
other solar activity---such as solar flares and coronal loops and mass
ejections. The physical driver of the magnetic activity cycle is the Sun's
magnetic field reversing polarity---that is, the Sun's north and south magnetic
poles swap places.

The beginning of the cycle is called the \emph{solar minimum}, and is marked by
observation of very few sunspots. Over time, observed solar activity increases
until it peaks at the \emph{solar maximum} midway through the cycle.
Conventionally, solar cycles are numbered from solar cycle 1, which began in
February 1755 and ended in June 1766. At the time of writing we are in solar
cycle 25, which began in December 2019. A full solar cycle from solar minimum
to solar minimum takes on average eleven years, meaning it takes roughly twenty
two years for the magnetic poles to return to their original positions. This
magnetic field periodicity is called the \emph{Hale cycle}.

(The magnetic fields of celestial bodies, such as stars and planets, are
theorized to be produced by the dynamo mechanism, which describes the motion of
plasma or molten metals moving within the body's core. The Earth's magnetic
field also undergoes polarity reversal. Though, unlike the Sun's solar cycle,
statistical analysis of the Earth's reversals have not revealed any obvious
periodicity \cite{Lutz85, Glatzmaiers95}.)

The solar cycle is the most obvious quasi-periodic cycle in the sunspot
sequence. However, there are other proposed periodic or quasi-periodic cycles
hypothesized to influence solar activity. The most relevant here is the
\emph{Gleissberg cycle}. This cycle's period is said to be on the order of $70$
to $100$ years and modulates the amplitude of the solar cycle. There are other
noncyclical observed patterns in the data. The Waldmeier effect states that the
length of an individual solar cycle is inversely proportional to its amplitude
maximum. The puzzling Gnevyshev-Ohl rule notes that sums of sunspot numbers
over odd cycles are highly correlated with sums over preceding even cycles. The
correlation is weaker between sums over even cycles and their preceding odd
cycles. This perhaps suggests a preferential labeling of the Hale cycle.
However, the physical mechanisms behind these patterns are still unknown. 

Let's examine the results of the \ekcs embedding. We set the past and future
length $\historylengthpast = \historylengthfuture$ to be twenty two years---the
length of the Hale cycle. The kernel bandwidth is set as the average amplitude
of a solar cycle. \Cref{fig:Sunspots-data} (right) shows the \ekcs embedded
using the first three components and colored according to the solar cycle
phase. This phase is estimated from data: after estimating the solar minima and
maxima with a low-pass filter, the approximate phase of the solar cycle for
each data point was computed by linear interpolation. Each \ekcs
$\ekcausalstate{\obvspast_t}$ was then colored according to the phase at time
$t$.

The \ekcs set $\ekCausalStateSet$ traces out a cone-like surface in
$\diffREigenvector$ space or, more fancifully, evokes a tornado. An immediate
observation when comparing to the $n$-butane example in
\cref{fig:n-butane-molecule} is how smooth the time ordering is in causal state
space. At this scale, we see no obvious time scale separation, indicating that
the distributions over futures change smoothly over time, rather than exhibit
sudden changes.

As is visually obvious from the color mapping, the first two components
$\diffREigenvector_1$ and $\diffREigenvector_2$ capture the eleven-year solar
cycle. Each eleven-year cycle traces out a roughly circular path in this plane
and the phase coloration approximately aligns cycle to cycle. This alignment is
not perfect because the solar cycle is not perfectly periodic. We can visually
identify phase slipping, as we expect for a system hypothesized to be modulated
by multiple, competing, periodic and quasi-periodic cycles. Yet,
\cref{fig:examples_eigenvalues} (center) shows eigenvalues $\diffEigenvalue_1$
and $\diffEigenvalue_2$ are almost identical, indicating no particular
preference for the phase of that cycle, unlike the crop example next.

The cycles are further organized along the third dimension
$\diffREigenvector_3$, here aligned with the vertical. Overlaying
$\diffREigenvector_3$ on the raw data, as in \cref{fig:Sunspots-data} (left),
shows that the third component captures the amplitude modulation of the solar
cycles with a slight offset---in fact, a lag of approximately eleven years.
When $\diffREigenvector_3$ is plotted with an eleven year shift, it tracks the
amplitude modulation of the solar cycle. This implies that the third component
at any given time is predicting the amplitude of the \emph{next} solar cycle,
capturing the patterns described the proposed Gleissberg cycle.
Additionally, an interesting section of the sunspot sequence is the
\emph{Dalton minimum}, which lasted from about 1790 to 1830 and covers solar
cycles 4 through 7. During this minimum the solar activity was unusually low.
In the \ekcs set the Dalton minimum corresponds to the segment of \ekcss that
traverse through the center of the ``cone'', indicating a significant change in
solar behavior as compared to other points in the sunspot series.

\begin{figure*}
\includegraphics[width=0.95\textwidth]{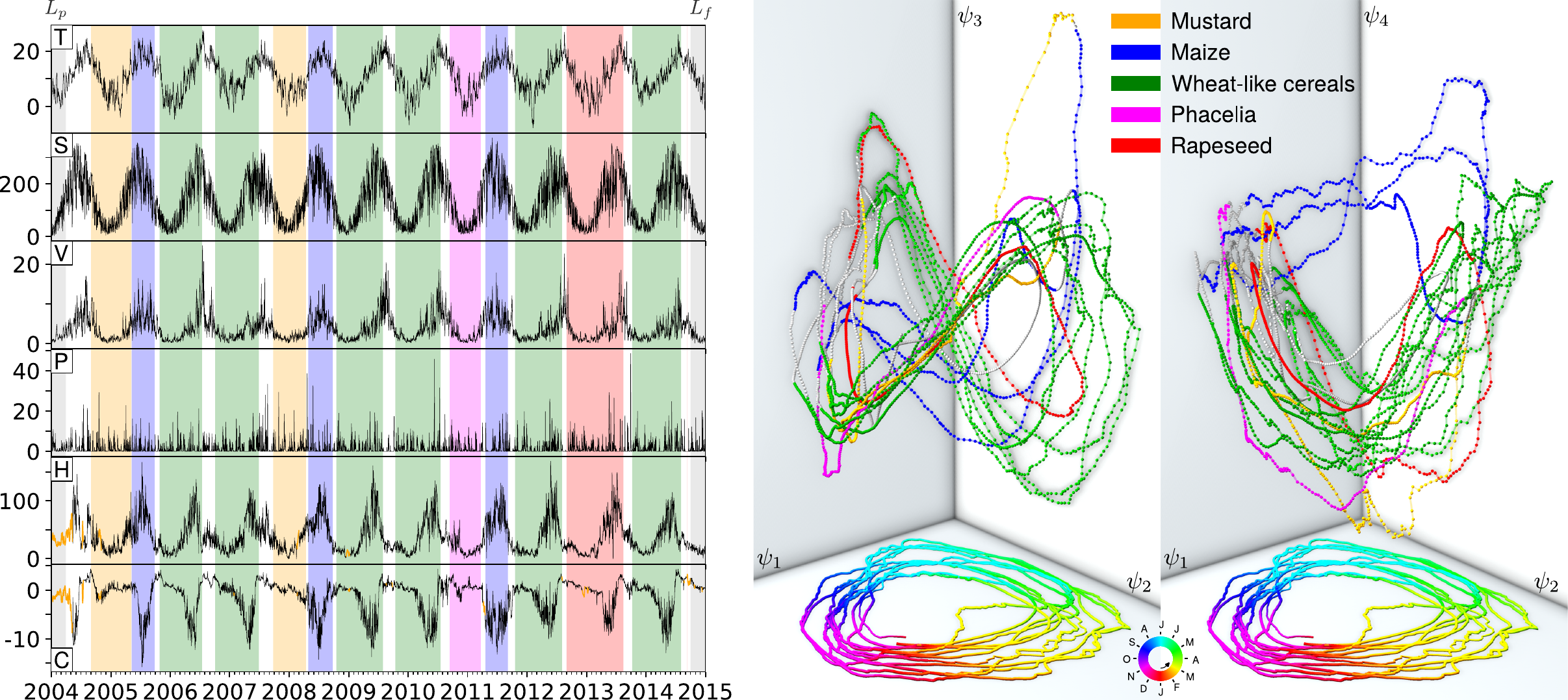}
\caption{Left: The six measurement sources over a period of 2004 to 2015 taken
	at the Grignon field site. From top to bottom: T) Temperature
	($\mathrm{^{\circ} C}$); S) Solar influx ($W / m^2$); V) Vapor pressure
	deficit ($\mathrm{hPa}$); P) Precipitation ($\mathrm{mm}$); H) Heat flux
	($\mathrm{W/m^{2}}$); C) $\mathrm{CO}_{2}$($\mathrm{\mu mol/m^{2}s}$).
	Missing or low-quality values that were filled with the gap-filling
	technique in \cref{app:gapfilling} are in orange. Periods between seedling
	and cutting/harvesting are also indicated for each culture type. Right: The
	\ekcss embedding on the left subplot is shown using the first three
	components $\left( \diffREigenvector_1, \diffREigenvector_2,
	\diffREigenvector_3 \right)$, and with $\diffREigenvector_4$ instead of
	$\diffREigenvector_3$ on the right. Trajectories are colored by culture
	type. Below, their projection on the
	$\left(\diffREigenvector_1,\diffREigenvector_2\right)$ plane is shown for
	both subplots, colored by the day of the year.
	}
\label{fig:crop_results}
\end{figure*}

\subsection{Grignon crop field}
\label{subsec:Crop-dynamics}

Our final example showcases how the \ekcs algorithm handles heterogeneous data
with a system of six different measurements from a crop field over a period of
eleven years. This example also highlights how to work with multivariate data
with periods of missing or low-quality measurements. These data are provided by
the Integrated Carbon Observation System (ICOS), a research collaboration
dedicated to producing standardized, high-quality climate observations to better
understand the carbon cycle and greenhouse gas balance in Europe and adjacent
regions. The collaboration maintains over forty atmospheric monitoring sites and
over one hundred ecosystem stations in over a dozen European countries. These
stations are strongly standardized and all observations are made publicly
available by the ICOS network \cite{ICOS_Gri}.

The Grignon ecosystem station is located in a nineteen hectare crop field about
forty kilometers west of Paris. The site is near a cattle farm and exposed to
heavy pollution from the Paris metropolitan area, as well as clean wind from the
southwest. The growing practices at the site are representative of standard
arable crop farming in France, with a rotation of crops including wheat, maize,
and oilseed rape.

The actual measurement device is an eddy covariance flux tower.  Eddy covariance
is the standard method used by ecosystem scientists to monitor gas exchange
between the land (soil and vegetation) and the atmosphere. The flux tower
measures the vertical movement of greenhouse gases carried by eddies, which over
time indicates whether the ecosystem is acting as a carbon sink or source. The
advantage of the eddy covariance method is that it measures fluxes directly
without disturbing the ecosystem. Flux towers also can continuously operate for
decades, providing a long-term picture of how an ecosystem is changing over
time. 

The tower at the Grignon site makes measurements of carbon fluxes and
meteorological conditions every half hour. We obtained measurements of air
temperature (\unit{\degreeCelsius}), solar influx (incoming shortwave radiation
\unit{W/m^2}), vapor pressure deficit (\unit{h\Pa}), precipitation (\unit{\mm}),
evapotranspiration (more precisely its proxy latent heat flux \unit{W/m^2}), and
the $\mathrm{CO}_{2}$ flux over the field (net ecosystem exchange
\unit{\umol/m^2s}) for a period covering 2004 to 2015. This six-dimensional data
set was measured daily over a period eleven years, totaling \num{4018} data
points. These measurements are plotted in \cref{fig:crop_results} (left),
overlaid with the crop type the field was growing at the time: either
mustard (yellow), maize (blue), cereals (green), Phacelia---a nitrogen holder
and weed suppressant (pink), and rapeseed (red). During the study period the
Grignon field grew cereals (wheat, barley, triticale) the most often.

Each measurement source is accompanied by a quality control flag indicating the
validity of the data at that time. In regions marked as low quality the data
may be irrelevant or missing entirely. This results in gaps in the time series
data in the latent heat flux and the $\mathrm{CO}_{2}$ flux series, highlighted
in orange in \cref{fig:crop_results} (left). We filled the missing values with
the technique outlined in \cref{app:gapfilling}. In brief, this method involves
training a linear model in \coords space, constrained by the valid measurement
series, and using the predictions of that model to fill the gaps. While simple,
this method was effective for our purposes and could be replaced by more
sophisticated gap-filling methods in the future.

Constructing kernels for heterogeneous data is discussed at length in
\cref{app:kernelMetaparameters} and so we will not go into depth here. However,
we do note that since all measurements are real valued the underlying metric
(Euclidean) was consistent across each measurement source. The past and future
history lengths of $\historylengthfuture = \historylengthpast = 91$ days (on
average, three months) were also held constant across measurement sources. The
Gaussian kernels bandwidth were set individually, as the standard deviation of
each source. The kernels were aggregated using the product rule with no
temporal decay.

\Cref{fig:crop_results} (right) presents two views of the \ekcss in diffusion
components space. The left subfigure uses $\diffREigenvector_3$ for the
vertical axis, while the right uses $\diffREigenvector_4$. Both
three-dimensional representations use $\diffREigenvector_1$ and
$\diffREigenvector_2$, which are also plotted in the plane below. This
projection is thus the same for both subfigures and it is colored according to
the day of year at time $t$ for each $\ekcausalstate{\obvspast_t}$. In both of
the three-dimensional plots, the \ekcss are colored according to the crop being
grown at their matching time. That said, it is important to keep in mind that
the $\ekcausalstate{\obvspast_t}$ are defined using the preceding three months
and predict the next three months of field dynamics
(\cref{eq:past_seq_def,eq:future_seq_def}).

We can visually identify that the first two components capture the seasonal
cycle. If we compare the phase alignment in this example to the phase alignment
in the sunspot $\ekCausalStateSet$ we can see that the alignment is a bit more
consistent, as would be expected for the regularity of the seasonal cycle
compared to the sunspots irregular 11-years cycle. However, the eigenvalues in
\cref{fig:examples_eigenvalues} (right) reveal that $\diffREigenvector_1$ is
more informative than $\diffREigenvector_2$ with respect to predicting the crop
dynamics. One possible interpretation is that fluctuations in the data sources
(temperature, rain, $\mathrm{CO}_{2}$, etc) in May or November (the alignment
of $\diffREigenvector_1$) are more impactful on the future of the system than
fluctuations in February or August (the alignment of $\diffREigenvector_2$).

Both views on the \ekcss reveal a saddle-like geometry in the three-dimensional
space. In the left subplot using $\diffREigenvector_3$, the low points of the
saddle are roughly aligned with the summer (green) and the winter (pink), while
the high points occur in fall (turquoise) and spring (yellow). In the
$\diffREigenvector_4$ subplot the saddle is inverted---the low points of the
saddle are aligned with fall and spring and the high points are aligned with
summer and winter. In late fall and early winter, the field will be bare for
the most of the prediction window (three months) and as such we see that the
\ekcss are mapped to a relatively small region in $\diffREigenvector$ space.

In the other seasons, the predictions made by the \ekcss begin to diverge. This
divergence begins to appear as early as February. At least some of this
divergence appears to be related to what pattern of crops is being grown that
year. Most years, the field grows only a single cereal, but in three years
(2005, 2008, and 2011) the field grows a cover crop in the spring (mustard in
2005 and 2008, Phacelia in 2011) followed by maize in the summer and fall. The
\ekcs associated with these years are mapped away from the body of the saddle
by both $\diffREigenvector_3$ and $\diffREigenvector_4$, although it is easier
to see with $\diffREigenvector_4$ in the right subplot. One possible
explanation for these diverging trajectories is that, given the same
environmental conditions, different plants produce different responses. This is
reflected in particular in the $\mathrm{CO}_{2}$ and latent heat series.

\section{Conclusion}
\label{sec:conclusion}

The development here covered a lot of ground---including, beyond the example
applications, reviewing the theory of computational mechanics
(\cref{sec:computuationalmechanics}) and its extension to kernel \eM introduced
in \cite{brodu22rkhs}. We then detailed the \ekcss algorithm
(\cref{sec:CausalStateEmbeddingAlgo}), including the diffusion map embedding it
uses (\cref{sec:diffMapping}) to produce \emph{\cdcs}. Finally, we gave four
examples of \ekcss computed on both simulated and real data, for both
deterministic and highly stochastic systems. We performed preliminary
structural analysis of these results, showing how the \cdcs uncover ``hidden''
structure. Three systems were arbitrarily high-dimensional, in the sense that
the number of degrees of freedom of the system was unknown a priori.

A major advantage of computational mechanics-based methods is the
interpretability the framework provides. By first transforming the sequences
into the \kcss space (\cref{sec:CausalStateEmbeddingAlgo}) and then applying
manifold-learning techniques (\cref{sec:diffMapping}), the \ekcss algorithm
provides a meaningful separation between the discovery of causal structure and
geometric structure. To the best of our knowledge the ability to analyze the
global geometry of the causal state set in this way is novel.

In a sense, for data analysis and modeling purposes, our \cdcs plays a similar
role as applying PCA to time-delayed embeddings. Both exploit the time
dependencies of the signal and perform dimension reduction. However, our \cdcs
are designed to maximize predictability, not variance, and they do so in a
principled way, with sound theoretical foundations
\cite{loomis2023topology,Uppe97a}. 

The \cdcs can be intuitively seen as capturing the most impactful factors
driving the system dynamics, as can be seen from the various examples recounted
here: recovery of the phase space for the simple pendulum
(\cref{subsec:Pendulum}), of the main energy wells and the structural
conformation of the butane molecule (\cref{subsec:Molecular-dynamics}), of the
major components of the solar cycle (phase, amplitude, \cref{subsec:Sunspots}),
of the seasonal cycle and the culture types in a cultivated crop
(\cref{subsec:Crop-dynamics}).

Finally, we emphasize the potential power of the full kernel \eM: it is not
only able to perform the kind of structural analysis done here, but also, given
sufficient data for its estimation, the kernel \eM is theoretically an
optimally predictive model of the system. These aspects require further
development and will be the topic of a future article in this series.

\section*{Authors' contributions}
\label{sec:acks}

Author Jurgens clarified and extended the theory of RKHS causal states and wrote the main part of the article. Author Brodu wrote the code, designed and ran the examples and wrote the first draft of the article. Both authors contributed equally to the analysis and the interpretation of the results.

\section*{Acknowledgments}
\label{sec:acks}

The authors thank Stefan Klus for providing the $n$-butane molecule simulation
data. We also thank the SILSO observatory for their helpful comments on the
role of the Gleissberg cycle in solar activity. In addition, we thank Jim
Crutchfield and Adam Rupe for helpful discussions and Jim for extensive
comments on the manuscript. This material is based on work supported by, or in
part by, the MELICERTES project (ANR-22-PEAE-0010) of the French National
Research Agency (France2030, national PEPR ``agro\'ecologie et num\'erique''
programmes), Inria's CONCAUST Exploratory Action, U.S. Army Research Laboratory
and U.S. Army Research Office Grant No. W911NF-21-1-0048, Templeton World
Charity Foundation (TWCF) grant TWCF0570, and Foundational Questions Institute
and Fetzer Franklin Fund grant FQXI-RFP-CPW-2007.

\bibliographystyle{ieeetr}
\bibliography{refsbrodu,ref,chaos}

\balance
\onecolumngrid
\vspace{2em}
\hrule
\vspace{2em}
\twocolumngrid

\appendix

\section{Code and data availability}
\label{app:codeanddata}

The code for reproducing these examples is freely available under the MIT
license from our project page:
\href{https://team.inria.fr/comcausa/continuous-causal-states/}
{https://team.inria.fr/comcausa/continuous-causal-states/}.
Data for the examples is also provided together with the source code.

\section{Metaparameters and kernel construction for sequences}
\label{app:kernelMetaparameters}

There are three categories of metaparameters:
\begin{itemize}
\item The specifics of the site-wise comparison, including kernel choices for
	each data source;
\item The method of kernel aggregation over time, including the temporal decay
	profile; and
\item The method of kernel aggregation over data sources, including potential
	weighting of the influence of each data source.
\end{itemize}
We review these below.

\subsection{Extended notations}
\label{app:extendedNotations}

\newcommand{\numTemporalBlocks} { K }
\newcommand{\tblockidx}          { k }

\Cref{sec:CausalStateEmbeddingAlgo} introduced notations for a single sequence
of observations and history lengths that do not depend on the data source. We
recall and extend these notations to more general cases, matching what is
actually handled by our algorithm:
\begin{itemize}
    \item We assume that $ \Obvs$ is constructed from measurements of $\numdatasources$ data sources: $\obvs_t = \left\{ \measurementsymbol^1_t,
    \measurementsymbol^2_t, \dots, \measurementsymbol^{\numdatasources}_t
    \right\}$. Each of these has its own data type: vector of real values,
	string, graph, and the like.

    \item For each data source, indexed by $\datasource=1\ldots \numdatasources$, the pasts are sequences of length $\historylengthpast^{\datasource}$ and the futures
    are sequences of length $\historylengthfuture^{\datasource}$. Specifically, they are the
    finite length sequences:
    \begin{align*}
        \obvspast^{\datasource}_{t} & = \left( \measurementsymbol^{\datasource}_{t - \historylengthpast^{\datasource} +1 },
        \dots, \measurementsymbol^{\datasource}_{t-1}, \measurementsymbol^{\datasource}_t \right) \quad \text{and} \\
        \obvsfuture^{\datasource}_{t} & = \left( \measurementsymbol^{\datasource}_{t+1}, \measurementsymbol^{\datasource}_{t +2},
        \dots, \measurementsymbol^{\datasource}_{t + \historylengthfuture^\datasource} \right)
    \end{align*}

    \item At each time, the pasts and futures are thus defined as lists of the above:
    \begin{align*}
        \obvspast_{t} & = \left( \obvspast^1_{t}, \ldots, \obvspast^{\numdatasources}_{t} \right) \quad \text{and} \\
        \obvsfuture_{t} & = \left( \obvsfuture^1_{t}, \ldots, \obvsfuture^{\numdatasources}_{t} \right)
    \end{align*}

    \item The data is organized in $\numTemporalBlocks$ temporal blocks of consecutive measurements. In \cref{sec:CausalStateEmbeddingAlgo} we considered only $\numTemporalBlocks=1$ block for simplicity, but the method can exploit measurements from multiple realizations of the same physical process. Each temporal block consists of $\totalLength_{\tblockidx}$ consecutive measurements $\left( \obvs_t, \ldots, \obvs_{t+\totalLength_{\tblockidx}-1} \right)$.

    \item With this data organization, have a library of $\numObvs = \sum_{\tblockidx} \left( \totalLength_{\tblockidx} - \max_\numdatasources{\historylengthpast^\datasource} -
    \max_\numdatasources{\historylengthfuture^\datasource} + 1 \right) $ pairs $\left( \obvspast_i, \obvsfuture_i \right)$ of pasts  $\ObvsPastSpace = \left\{ \obvspast_1, \dots, \obvspast_{\numObvs} \right\}$ and futures
    $\ObvsFutureSpace = \left\{ \obvsfuture_1, \dots,
    \obvsfuture_{\numObvs} \right\}$. Each of the $i=1\dots \numObvs$ pairs corresponds to a time index in one of the temporal blocks.

    \item Some data may be missing or be tagged as low quality. These are replaced by \textsf{not-a-number} pseudo-values and possibly gap-filled with the technique described in \cref{app:gapfilling}. Otherwise, each of these missing data reduces $\numObvs$ by at most $2\times\left( \historylengthpast^{\datasource} +
    \historylengthfuture^{\datasource}\right) - 1$ values (consecutive missing data reduce by less than this amount).

\end{itemize}

The next subsections describe how to construct the two symmetric and positive definite reproducing kernels
$\kernelPast \left( \obvspast, \cdot \right) : \ObvsPastSpace \to
\HilbertSpacePast $ and $\kernelFuture \left( \obvsfuture, \cdot \right) :
\ObvsFutureSpace \to \HilbertSpaceFuture$, for pasts and futures respectively. These kernels generate
the reproducing kernel Hilbert spaces $\HilbertSpacePast$ and
$\HilbertSpaceFuture$.

\subsection{Sitewise comparison}
\label{app:sitewiseComparison}

As already noted, we do not constrain our measurements $\measurementsymbol^i$ to
be anything other than drawn from a compact set. Typically, this is either a
finite discrete set or a closed interval in the reals. This is because a
\emph{universal} reproducing kernel over a compact set has several desirable
properties, namely that the measure embedding into the reproducing Hilbert space
generated by the kernel given by \cref{eq:RKHSEmebddingFunction} is injective
and the norm convergence under the kernel inner product as defined in
\cref{eq:RKHSInnerProduct} is equivalent to convergence in distribution over
measures \cite{Sriperumbudur2010}.

A very large class of universal kernels are those which are both compactly supported
and translation invariant in $\reals^n$ \cite{sriperumbudur2010universality}.
This includes popular kernels such as the Gaussian and the Laplacian, that both
depend only on the distance between their arguments. Typically, the distance is
also scaled by a kernel width or radius that sets the characteristic scale.
For instance, the Gaussian kernel is written:
\begin{align*}
    \kernel_G \left( \measurementsymbol, \measurementsymbol' \right) =
    \exp \left( - \frac{1}{2} \left (\| \measurementsymbol - \measurementsymbol' \| / \kernelBandwidth \right)^2 \right).
\end{align*}

So, it is possible to define a kernel for a particular data source
$\kernel^{\datasource} \left( \measurementsymbol^{\datasource}, \measurementsymbol'^{\datasource} \right)$ simply by defining a metric on that data source $\datasource$
and picking an appropriate bandwidth $\kernelBandwidth$, expressed in data units.

The choice of metric is fundamental to the \ekcss construction because
it sets the underlying geometry that the kernel uses to embed the causal
states, so it must be done consistently with the nature of the underlying system.
However, we have considerable freedom in picking $\| \measurementsymbol - \measurementsymbol' \|$: it only needs to be a metric. Since metrics have been defined on all
kinds of mathematical structures---strings, graphs, probability distributions,
game theoretic strategies---our ability to build kernels is extraordinarily
broad. In practice, a Euclidean metric is typically used when the underlying data is
drawn from $\reals^n$ and a discrete metric is used for symbolic data. In this
paper examples, all the data sources are real numbers and we use the Euclidean
metric together with the Gaussian kernel.

With these choices fixed, the most important parameter becomes the kernel
bandwidth $\kernelBandwidth$, as it sets the scale for comparing two data values.
For example, exploring the effect of temperature variations on the cultures at the scale of $0.1\mathrm{^{\circ} C}$
would make no sense for the crop example in \cref{subsec:Crop-dynamics}. When a natural, characteristic scale exists
from domain knowledge, it should be used to analyze the physical process operating at that scale.
In a data exploration phase, it is also interesting to sweep through a range of bandwidths. This could highlight
structure at multiple scales, possibly reflecting separate physical processes.

\subsection{Kernel aggregation over time}
\label{app:kernelAggregationTime}

Once kernels have been set to compare values for each data source, the next step is to combine them across time to get kernels over sequences.

Fortunately, kernels can be combined easily. Let $\kernel^A \left(a, a' \right) : A \times A \to \reals$ and $\kernel^B
\left( b ,b' \right) : B \times B \to \reals$  be reproducing kernels. Then the
following operations result in a reproducing kernel $\kernel^C$:
\begin{itemize}

    \item Scalar multiplication by $\alpha >0$ : $ \kernel^C \left( a,a' \right)
    = \alpha \kernel^A \left( a,a' \right) $, where $C = A$.

    \item Exponentiation by $\alpha >0$: $ \kernel^C \left( a,a' \right) = \left(
    \kernel^A \left( a,a' \right) \right)^{\alpha}$, where $C = A$.

    \item Kernel multiplication: $\kernel^C \left(c, c' \right) = \kernel^A \left(
    a,a' \right) \kernel^B \left( b, b' \right)$, where $ C = A \times B$.

    \item Kernel addition: $\kernel^C \left(c, c' \right) = \kernel^A \left(
    a,a' \right) + \kernel^B \left( b, b' \right)$, where $ C = A \times B$.
\end{itemize}

The above relations are not exhaustive (see \cite[Section 6.2]{bishop2006pattern}) but
they are enough to cover the most usual cases. They also preserve the translation-invariance
property, hence guarantee the universality of the combined kernels (with our compactness requirement).

Kernels over sequences can be then easily defined by combining the kernels across time, using schemes such as:
\begin{itemize}
    \item Geometric weighting scheme:
    \begin{align*}
        \kernel^{\ObvsFuture^{\datasource}} \left( \obvsfuture^{\datasource}_t, \cdot \right) =
        \prod_{\tau=1}^{\historylengthfuture^\datasource}
        \kernel^{\datasource}
        \left( \measurementsymbol^{\datasource}_{t+\tau},
        \cdot \right)^{\omega\left(\tau\right)}
    \end{align*}
    \item Arithmetic weighting scheme:
    \begin{align*}
        \kernel^{\ObvsFuture^{\datasource}} \left( \obvsfuture^{\datasource}_t, \cdot \right) =
        \sum_{\tau=1}^{\historylengthfuture^\datasource}
        \omega\left(\tau\right) \kernel^{\datasource}
        \left( \measurementsymbol^{\datasource}_{t+\tau},
        \cdot \right)
    \end{align*}
\end{itemize}

where $\kernel^{\ObvsFuture^{\datasource}}$ is the kernel over the future sequences $\ObvsFuture^{\datasource}$ of data source $\datasource$. The kernel can be further normalized so that $\kernel^{\ObvsFuture^{\datasource}} \left( \obvsfuture^{\datasource}_t, \obvsfuture^{\datasource}_t \right) = 1$. Similar kernel definitions hold for the past sequences $\ObvsPast^{\datasource}$ of each data source.

In these schemes, we call $\omega\left(\tau\right)$ the decay profile. Typically, we desire a weighting scheme designed such that observations further away in time have less influence than the more immediate past/future observations. In practice, when $\| \measurementsymbol^{\datasource} - \measurementsymbol'^{\datasource} \|$ is the Euclidean metric, and when $\kernel^{\datasource}$ is the Gaussian kernel, then using the product scheme becomes a natural choice:
\begin{align*}
    \kernel^{\ObvsFuture^{\datasource}} \left( \obvsfuture^{\datasource}_t, \obvsfuture'^{\datasource}_t \right) =
    \prod_{\tau=1}^{\historylengthfuture^\datasource}
    \mathrm{exp}\left(-\frac{1}{2\xi^2}
    \| \measurementsymbol^{\datasource}_{t+\tau} - \measurementsymbol'^{\datasource}_{t+\tau}\|^2
    \right)^{\omega\left(\tau\right)} \\
    =\mathrm{exp}\left(-\frac{1}{2\xi^2}
    \sum_{\tau=1}^{\historylengthfuture^\datasource} \omega\left(\tau\right) \| \measurementsymbol^{\datasource}_{t+\tau} - \measurementsymbol'^{\datasource}_{t+\tau}\|^2
    \right)
\end{align*}

Then, the weighting scheme acts on the distance between measurements. It can be interpreted as scaling the bandwidth $\xi$ by $1/\sqrt{\omega\left(\tau\right)}$ for each $\tau$, hence reducing the sensitivity of the comparisons as $\omega$ decays.

Our previous work \cite{brodu22rkhs} introduced a power law decay. Exponential decay was successfully used in a related work \cite{rupe2019disco}, where it was noted to give superior performances compared to uniform weighting. An appealing avenue is to tie the weights to the autocorrelation of each data source. An information-theoretic based method was also proposed in
\cite{loomis2023topology}. Since the only constraint is that the aggregation of kernels be bounded, there is considerable flexibility in how one wishes to define $\omega\left(\tau\right)$. Mathematically, we must introduce some kind weighting scheme on the sums of kernels acting on infinite sequences that ensures their convergence \cite{loomis2023topology}. A temporal decay fills this role naturally. Practically, however, we work with finite sequences and so a decay profile is not strictly mathematically necessary.

A common misconception is that without a decay the aggregated kernel is invariant under permutations of the sequence, and that would make it inappropriate for use on time series. While the resulting kernel would indeed be invariant, this is the expected behavior and not a problem, even when working with time series. Choosing not to use a decay profile is, in effect, making the assumption that the system itself does not experience causality decay (or that the causality decay is not appreciable within the analyzed time window), in which
case it is perfectly acceptable for the distance between $\obvspast_{1}$ and $\obvspast_{2}$ to be the same as the distance between their permuted versions $\obvspast'_{1}$ and $\obvspast'_{2}$. This is not, in itself, a reason to select a causal decay profile which enhances preferentially certain permutations over others.

In the examples here, we only use the product scheme with uniform
weighting---that is, without temporal decay. This choice was made to reduce the
number of metaparameters to set, also to ease comparison between examples.

\subsection{Kernel aggregation over data sources}
\label{app:kernelAggregationSources}

Once the kernels $\kernel^{\ObvsFuture^{\datasource}}$ are constructed for each data source, the procedure for combining them is similar to the aggregation over time. For example, using the arithmetic scheme:
\begin{align*}
    \kernel^{\ObvsFuture} \left( \obvsfuture_t, \cdot \right) =
    \sum_{\datasource=1}^{\numdatasources}
    w\left(\datasource\right) \kernel^{\ObvsFuture^{\datasource}}
    \left( \obvsfuture^{\datasource}_t, \cdot \right)
\end{align*}

and similarly for the geometric scheme. The weighting profile $w\left(\datasource\right)$ now sets the influence of each data source in the final, combined kernel. This may be useful for expressing domain knowledge, but otherwise uniform weighting should be considered. In principle, it would be possible to use a different weighting scheme for the kernel over futures $\kernel^{\ObvsFuture}$ and for the kernel over pasts $\kernel^{\ObvsPast}$, which is built with the same procedure. Again, there is no motivation to do so without prior knowledge on the data.

\section{Gap-filling missing data}
\label{app:gapfilling}

When working with real measurements, there are sometimes sections with missing or invalid
data. For our method, these missing measurements can be quite impactful: because
of the way we construct libraries of past and future sequences, even a single
missing (or invalid) measurement $\measurementsymbol_{t}^{\datasource}$ in data
source $\datasource$ at time $t$ induces a gap length of $\gapLength =
\times\left( \historylengthpast^{\datasource} +
\historylengthfuture^{\datasource} \right) - 1$. This because before the
missing measurement, the last future sequence that can be computed is
$\obvsfuture^{\datasource}_{t-\historylengthfuture^{\datasource}-1}$ and after the gap, the first past history
that can be computed is $\obvspast^{\datasource}_{t+\historylengthpast^{\datasource}}$. Consecutive
missing values, however, only increase $G$ by \num{1} for each additional missing
value.

\subsection{Linear interpolation of causal diffusion coordinates}

Filling these gaps amounts to predicting the missing values. As a proof of
concept, we train a linear transition operator $\gapfillingTransOp$ acting on the causal
diffusion coordinates:
\begin{align}
    \label{eq:linear_transition_op}
    \left[ \diffcoord{1}{t+1}, \ldots, \diffcoord{M}{t+1}\right]^{\intercal} \: \hat{=} \: \gapfillingTransOp \left[\diffcoord{1}{t}, \ldots, \diffcoord{M}{t}\right]^{\intercal}
\end{align}
$\gapfillingTransOp$ is typically fit by minimizing the least squared error for
\cref{eq:linear_transition_op} using all valid time transitions
$\ekcausalstate{\obvspast_t} \to \ekcausalstate{\obvspast_{t+1}}$.

For each data gap of size $\gapLength$, consider the time $t$
for the last valid \ekcs $\ekcausalstate{t}$ before this gap.
The next available state is then $\ekcausalstate{t + \gapLength+1}$.
With a slight abuse of notation, remembering that $\gapfillingTransOp$ acts on the diffusion coordinates, we fill the gaps by applying $\gapfillingTransOp$ both forward:
\begin{align*}
    \ekcausalstate{}_{t+g}^f = \gapfillingTransOp^g \: \ekcausalstate{t}
\end{align*}
and backward:
\begin{align*}
    \ekcausalstate{}_{t+g}^b = \gapfillingTransOp^{g-\gapLength-1} \: \ekcausalstate{t + \gapLength+1}
\end{align*}
then combining both:
\begin{align}
    \ekcausalstate{}_{t+g}^p = \left( 1 - \frac{g}{\gapLength+1} \right) \ekcausalstate{}_{t+g}^f + \frac{g}{\gapLength+1} \ekcausalstate{}_{t+g}^b
    \label{eq:gap_filling_combi}
  ~.
\end{align}
This way, we get predicted states $\ekcausalstate{}_{t+g}^p$ that recover the valid states before ($g=0$) and after ($g=\gapLength+1$) the gap and interpolate between these within the gap ($1\leq g \leq \gapLength$).

Once intermediate states are predicted, we still need to convert them into predictions in data space, for each data source. For this gap-filling proof of concept, we also fit these observation functions $\obsfun^d$ as linear maps, using all valid measurements:
\begin{align*}
    \measurementsymbol^\datasource_t \: \hat{=} \: \obsfun^d \left[ \diffcoord{1}{t}, \ldots, \diffcoord{M}{t}\right]^{\intercal}
  ~.
\end{align*}
In other words, each $\obsfun^d$ attempts to recover the corresponding data source value from the diffusion coordinates of the \ekcs matching that time. We make predictions for all missing values by applying these $\obsfun^d$ maps to the intermediate states $\ekcausalstate{}_{t+g}^p$ that were interpolated above.

The linear maps $\obsfun^d$ and operator $\gapfillingTransOp$ do not reflect
the correct \eM dynamic. A full treatment would consider the possible
distribution of intermediate states, given the observed values before and after
the gap, and produce an ensemble of intermediate trajectories, from which data
samples could be drawn. Still, a linear interpolation and prediction model is
sufficient our purposes here.

\subsection{Using all valid measurements}

It is common that only a subset of the data sources have missing values, such as the latent heat and the $\mathrm{CO}_2$ flux in the example in \cref{subsec:Crop-dynamics}. The other data sources present valid measurements at the same time indices, which shall be exploited to refine the gap filling.

Consider the forward computations $\ekcausalstate{}_{t+1}^f = \gapfillingTransOp \: \ekcausalstate{t}$. After each application of $\gapfillingTransOp$, it is possible to apply the observation functions $\obsfun^d$ on the predicted state $\ekcausalstate{}_{t+1}^f$, yielding a vector $\left( \hat{\measurementsymbol}^1_{t+1}, \ldots, \hat{\measurementsymbol}^\numdatasources_{t+1} \right)$ of $\numdatasources$ measurement estimates. For the valid data sources, the measurement estimates and the actual data should ideally match. Noting $V \subset \left\{ 1\ldots \numdatasources \right\} $ the set of valid data source indices at time $t+1$, we implement the following optimization scheme:
\begin{align}
    \tilde{\causalstate}^f_{t+1} = &
    \textrm{argmin}_{s} \,
    \sum_{\datasource\in V} w\left(\datasource\right) \left( \obsfun^\datasource s -
    \measurementsymbol^\datasource_{t+1} \right)^2 \nonumber \\
    \textrm{s.t. } & \left\Vert s - \ekcausalstate{}_{t+1}^f \right\Vert
    <\epsilon \left\Vert \ekcausalstate{}_{t+1}^f \right\Vert
    \label{eq:gap_filling_opti}
  ~,
\end{align}
using the CVXPY iterative constrained convex optimizer with starting point
$\causalstate_{t+1}^{f}$. In other words, we seek a causal state estimate
$\tilde{\causalstate}^f_{t+1}$ maximally consistent with the valid data
$\left\{ \measurementsymbol^\datasource_{t+1} \right\}_{\datasource\in V}$,
but not too far from the forward state estimate $\ekcausalstate{}_{t+1}^f$
computed with the transition operator $\gapfillingTransOp$.

\begin{figure*}[t]
    \includegraphics[width=1\textwidth]{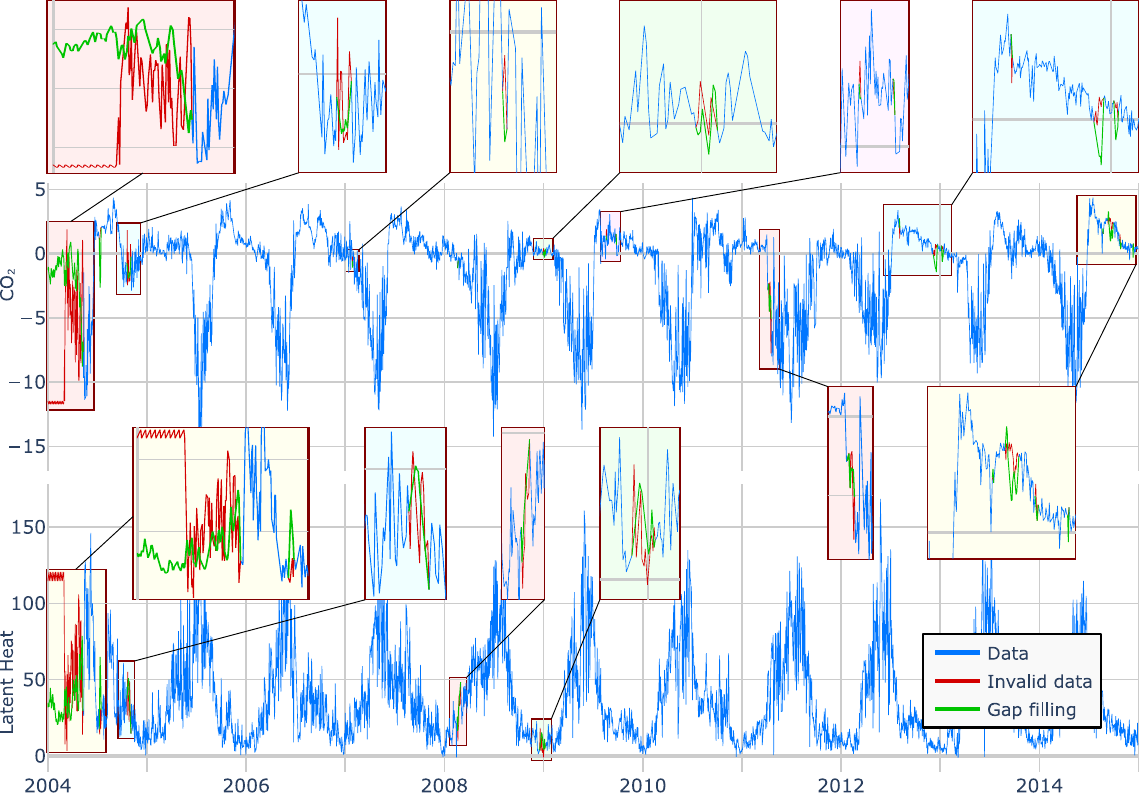}
    \caption{ $\mathrm{CO}_{2}$ and Latent Heat flux data from the crop example in \cref{subsec:Crop-dynamics}.
        Data in red is marked as invalid and should not be seen as a target for
        the gap-filled data (in green). This is especially apparent for the first
        few months of 2004. Data in blue could also be of low quality,
        especially at each gap boundary. The gap-filled sections look plausible
        and are consistent with the measurements of all other variables. }
    \label{fig:gap_filling_details}
\end{figure*}

This procedure is repeated iteratively $\gapLength$ times in the forward
direction: $\ekcausalstate{}_{t+g+1}^f = \gapfillingTransOp \:
\tilde{\causalstate}^f_{t+g}$ is used as a starting point for optimizing
$\tilde{\causalstate}^f_{t+g+1}$ in each of these steps.

The same procedure is applied backward and the final predicted states
$\tilde{\causalstate}^p_{t+g}$, for all indices $g$ in the gap, are combined
using \cref{eq:gap_filling_combi}. Finally, predicted values
$\hat{\measurementsymbol}^\datasource_{t+g} = \obsfun^\datasource \: \tilde{\causalstate}^p_{t+g}$ are set for each missing data
source $\datasource$ at each time index $t+g$ in the gap.

\subsection{Adjusting the \cdcs}

By construction the predicted states for the gaps respect the constraint
$\left\Vert s - \ekcausalstate{}_{t+g}^f  \right\Vert
< \epsilon \left\Vert \ekcausalstate{}_{t+g}^f \right\Vert $ of
\cref{eq:gap_filling_opti} at each step of the iterative process, and similarly
for the backward pass. Errors may accumulate but, at most, only proportionally to a fixed number of
iterations $\gapLength$. The predicted state $\tilde{\causalstate}^p_{t+g}$ thus remains arbitrarily close to the
$\ekcausalstate{}_{t+g}^p$ of \cref{eq:gap_filling_combi}. These were computed
with the linear transition operator $\gapfillingTransOp$, itself fit by a linear
regression. Hence, the rank of the original $\numObvs \times \numObvs$
similarity matrix $\gramMatCS_{ij} =
\innerProductMath{\ekcausalstate{\obvspast_i}}{\ekcausalstate{\obvspast_j}}$
should not be changed by the addition of the predicted states. The updated
$\widetilde{\gramMatCS}$ is now a similarity matrix of size
$\left(\numObvs+\gapLength\right) \times \left(\numObvs+\gapLength\right)$
between all states, original and produced to fill the gap.

We take the conservative approach to first use only the gap-filled states for the $\max_\numdatasources \historylengthpast^\datasource + \historylengthfuture^\datasource - 1$ states before and after the missing values. Using only the backward or the forward pass, we also fill the sections with partial histories at the beginning and the end of each of the $\numTemporalBlocks$ temporal blocks. In other words, the optimization in \cref{eq:gap_filling_opti} was done only for time indices that match actual data, with no missing value, but for which an \ekcs could not be estimated due to lack of full histories. After this first step, the gaps consist only of the time indices with actually missing data. We recompute the diffusion transform on the updated $\widetilde{\gramMatCS}$. The whole gap-filling procedure is then repeated in this updated \cdcs space, for filling these much smaller gaps. As a result, we get state estimates for every of the original data time indices, using both the original $\ekcausalstate{}$ embeddings and the predicted $\tilde{\causalstate}^p$ for partial histories and missing data.

\Cref{fig:gap_filling_details} shows the data for the $\mathrm{CO}_{2}$ flux over
the whole series, together with zooms on gap-filled sections. The sections
labeled “invalid data” (in red) actually correspond to a quality control flag of
$0$, meaning these data are not reliable at all and should be discarded. Hence,
the invalid data in \cref{fig:gap_filling_details} should not be seen as a target
for the gap-filled data (in green). The valid (blue) data could also be of low
quality at each gap boundary. What is relevant is that gap-filled data in
\cref{fig:gap_filling_details} look plausible and could have been measured instead
of the invalid red values. The gap-filled data indeed look like they could have
been measured, even for the first few months of 2004 for which the invalid data
is clearly nonsensical.

In this proof of concept for gap-filling, the transition operator $\gapfillingTransOp$ and the observation functions $\obsfun^\datasource$ are linear, but they could be replaced by full-fledged estimators of the \keM dynamics in future works. Compared to alternative gap-filling techniques, this method presents the advantage of exploiting all available information coming from all data sources at all time indices, and consistently so with the overall system dynamics.

\end{document}